\pdfoutput=1
\documentclass[11pt]{article}

\usepackage[final]{acl}

\usepackage{times}
\usepackage{latexsym}

\usepackage[T1]{fontenc}
\usepackage[utf8]{inputenc}

\usepackage{microtype}

\usepackage{inconsolata}
\usepackage{microtype}
\usepackage{graphicx}
\usepackage{subfigure}
\usepackage{booktabs} % for professional tables

\usepackage{hyperref}
\usepackage{graphicx}
\usepackage{amsmath}
\usepackage{amssymb}
\usepackage{mathtools}
\usepackage{amsthm}
\usepackage{amsfonts} 
\usepackage{nicefrac}  

\usepackage{xcolor}         % colors
\usepackage[export]{adjustbox}
\usepackage{wrapfig}
\usepackage{multirow}
\usepackage{CJKutf8}

\usepackage[capitalize,noabbrev]{cleveref}
\title{\includegraphics[scale=0.04,valign=c]{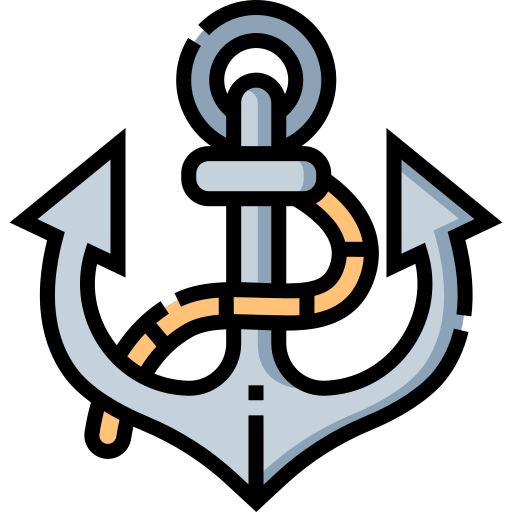} Anchored Answers: Unravelling Positional Bias \\ in GPT-2's Multiple-Choice Questions}

\author{
 \textbf{Ruizhe Li\textsuperscript{1}},
 \textbf{Yanjun Gao\textsuperscript{2}}
\\
 \textsuperscript{1}Department of Computing Science, University of Aberdeen, UK\\
 \textsuperscript{2}University of Colorado Anschutz Medical Campus, United States
\\
 \small{
   \textbf{Correspondence:} \href{mailto:ruizhe.li@abdn.ac.uk}{ruizhe.li@abdn.ac.uk}
 }
}

\begin{document}
\maketitle
\begin{abstract}
Large Language Models (LLMs), such as the GPT-4 and LLaMA families, have demonstrated considerable success across diverse tasks, including multiple-choice questions (MCQs). However, these models exhibit a positional bias, particularly an even worse ``\textbf{anchored bias}'' in the GPT-2 family, where they consistently favour the first choice `\texttt{A}' in MCQs during inference. This anchored bias challenges the integrity of GPT-2's decision-making process, as it skews performance based on the position rather than the content of the choices in MCQs. In this study, we utilise the mechanistic interpretability approach to identify the internal modules within GPT-2 models responsible for this bias. We focus on the Multi-Layer Perceptron (MLP) layers and attention heads, using the ``logit lens'' method to trace and modify the specific value vectors that contribute to the bias. By updating these vectors within MLP and recalibrating attention patterns to neutralise the preference for the first choice `\texttt{A}', we effectively mitigate the anchored bias. Our interventions not only mitigate the bias but also improve the overall MCQ prediction accuracy for the GPT-2 family across various datasets. This work represents the first comprehensive mechanistic analysis of anchored bias from the failing cases in MCQs within the GPT-2 models, introducing targeted, minimal-intervention strategies that significantly enhance GPT2 model robustness and accuracy in MCQs.
Our code is available at \url{https://github.com/ruizheliUOA/Anchored_Bias_GPT2}.
\end{abstract}

\section{Introduction}

% \vspace{-1em}
Large Language Models (LLMs) exhibit remarkable capabilities across a wide array of tasks, including multiple-choice question (MCQ)~\citep{robinson2022leveraging}, which are largely attributed to the advancements in the Transformer backbone. These models not only excel at reasoning but also demonstrate significant inductive capabilities, which make them highly effective in different domains~\citep{hu-etal-2023-hearing,chen2023meditron,hu-etal-2023-mir,team2023gemini,anil2023palm,hu-etal-2024-listen,hu-etal-2024-gentranslate,li2025attributing,liu2025towards,ji2024detecting,info:doi/10.2196/58670}.

\begin{figure}[tb]
    \begin{center}
    \includegraphics[width=0.99\columnwidth]{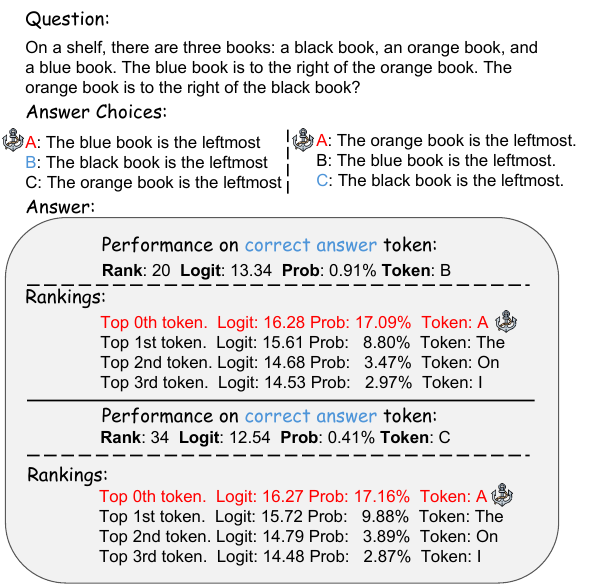}
    \end{center}
    % \vspace{-1em}
    \caption{MCQ prompt paradigm used in GPT2-Small and next token logit rankings with probability during inference. Regardless of the order in which correct answer choices are placed in the prompt except `\texttt{A}', GPT2-Small always give a higher logit score to the choice immediately following the \texttt{Answer Choices:}, i.e., \textcolor{red}{\texttt{A}}, where \includegraphics[scale=0.03,valign=c]{figures/anchor.png} represents the anchored bias for the incorrect choices (the correct choices should be \textcolor{blue}{\texttt{B}} and \textcolor{blue}{\texttt{C}} for this example).}
    % \vspace{-5mm}
    \label{fig:MCQ_prompt}
\end{figure}

Despite their success, recent studies have uncovered a notable flaw: LLMs exhibit a positional bias when tasked with MCQs. Specifically, the performance of these models (e.g., LLaMA~\citep{touvron2023llama}, LLaMA2~\citep{touvron2023llama2}, GPT-4~\citep{achiam2023gpt}) varies significantly depending on the position of the correct answer within the given choices~\citep{pezeshkpour2023large,zheng2023large}. We further observe that this vulnerability to positional bias is even worse in the GPT-2 family, ranging from GPT2-Small-124M to GPT2-XL-1.5B~\citep{radford2019language}. Our investigations reveal that GPT-2 models consistently favour the first choice `\texttt{A}', regardless of the actual position in the input MCQ prompt where the correct answer choice is placed, which we term as ``\textbf{anchored bias}'' in Fig.~\ref{fig:MCQ_prompt}.

Previous work primarily mitigated positional bias in MCQ by analysing the impact of different prompt structures~\citep{pezeshkpour2023large} or by estimating different datasets' prior bias based on test samples~\citep{zheng2023large}. Such approaches often remain superficial, merely altering the prompt presentation, or lacking a comprehensive analysis of fundamental reasons. While~\citet{lieberum2023does,wiegreffe2025answer} investigated positional bias-related problem using mechanistic interpretability, they mainly focus on success cases. There has been a lack of investigation into the internal mechanisms of LLMs from the failing cases that contribute to the \textbf{anchored bias} and strategies to mitigate it without the need for prompt engineering or prior estimation.

We apply mechanistic interpretability to reverse-engineer the internal workings of the GPT-2 family to understand the origins and extent of the anchored bias in the failing cases. We quantitatively demonstrate that the GPT-2 Small, Medium, Large, and XL models exhibit this anchored bias with significant regularity across various MCQ datasets, ranging from 2-choice to 5-choice settings. Our detailed analysis using the ``logit lens''~\citep{logitlens2020} approach localises Multi-Layer Perceptron (MLP) layers with specific dimensionality and attention heads that disproportionately influence this anchored bias. We find that certain value vectors in the MLP, which inherently harbour this bias, and specific attention heads pay more weight on the `\texttt{A}' position over the correct answer choice positions in the input prompt.

Inspired from~\citep{geva-etal-2021-transformer,geva-etal-2022-transformer} where MLPs can be treated as key-value memories, we use a straightforward yet potent method~\citep{dai-etal-2022-knowledge} to update these critical value vectors in the MLP, effectively mitigating the anchored bias. This adjustment not only mitigates the anchored bias but also enhances the overall MCQ prediction accuracy over 70\% averaged across various MCQ datasets and all GPT-2 family. Additionally, we propose a novel strategy to recalibrate the attention patterns by swapping the attention weight between the anchored position and the correct answer choice position. This strategy also mitigates the anchored bias to a certain extent, especially for the classification accuracy improvement of the Indirect Object Identification (IOI) dataset~\citep{wang2022interpretability} over 90\% on GPT2-Medium. Finally, we trace the full anchored bias circuit of each GPT2 model, which includes all attention heads and MLPs contributing to this bias.

In conclusion, to the best of our knowledge, this work is the first comprehensive mechanistic analysis of the intrinsic anchored bias from the failing cases in MCQ tasks across the entire GPT-2 family. By identifying and rectifying the critical value vectors within MLP and attention heads responsible for this bias, we introduce novel, minimal-intervention strategies that significantly reduce GPT-2 models' vulnerability and enhance robustness against anchored bias in the MCQ task.
% \vspace{-1.5mm}
\section{Related Work}

Several studies have documented the effects of positional bias on LLM accuracy in MCQs. \citet{pezeshkpour2023large} found a "sensitivity gap" in models like GPT-4, where positional bias can decrease performance by up to 75\% in a zero-shot setting, and they improved accuracy with new calibration strategies. \citet{wang2023large} also noted the impact of option order on GPT-4’s scores, enhancing accuracy through a calibration framework including multiple evidence and balanced position adjustments, along with human involvement. \citet{zheng2023large} addressed "selection bias" where LLMs disproportionately favour certain options, introducing a debiasing method, PriDe, that adjusts predictions during inference. \citet{wang2024beyond} explored performance changes from reordering answer options, confirming that this impacts understanding. \citet{turpin2024language} and \citet{zheng2024judging} explore the effects of positional bias in LLMs, showing how it skews Chain-of-Thought generation and evaluator judgments, and emphasise the need for strategies to detect and mitigate these biases. However, those studies did not analyse such bias within models and further identify which component is relevant. Recently, ~\citet{lieberum2023does} analyses final-token attention heads and identifies a subset ``correct letter heads'', which focus on earlier answer symbols to promote the correct choice based on its order (mainly for A/B/C/D). However, their findings are based solely on MMLU task using the closed-source Chinchilla-70B.~\citet{wiegreffe2025answer} investigates how successful models perform formatted MCQs with symbol binding internally rather than focusing on the failure cases when models have positional bias. Compared to~\citep{lieberum2023does,wiegreffe2025answer} focusing on the success cases, our work is focused on the failing cases, and we believe that these complementary perspectives help form a more complete understanding of how LLMs handle MCQs.

\section{Background: Large Language Models and Mechanistic Interpretability}\label{s:background}
% \vspace{-1mm}
\begin{figure*}[tb]
% \vspace{-3mm}
    \centering
    \includegraphics[scale=0.45]{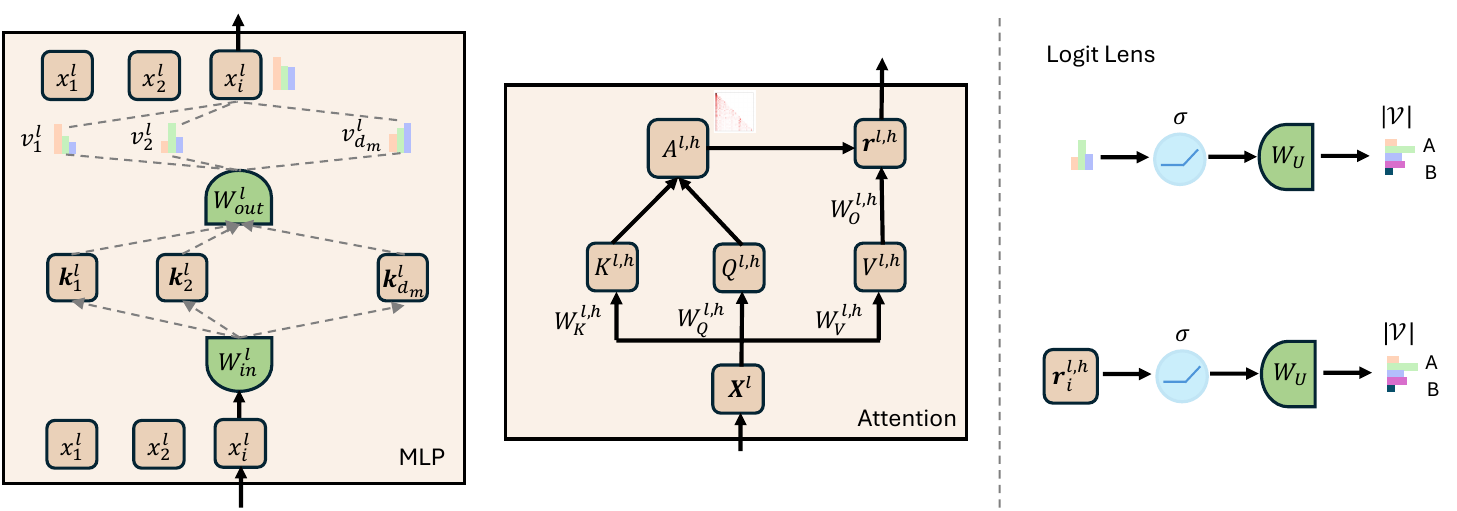}
    % \vspace{-3mm}
    \caption{\textit{Left}: the MLP and attention modules of LLMs, where the input prompt is encoded via $W_E$, and then the processed information via attention and MLP layer is accumulated back to the residual stream  $\mathbf{X}^\ell$ at layer $\ell$. Finally, the residual stream at $L$ layer is unembedded as logits and normalised as a probability distribution for next token prediction. \textit{Right}: logit lens~\citep{logitlens2020} is used to investigate the contribution of attention pattern and MLP module for the next token prediction.}
    \label{fig:transformer}
    % \vspace{-1.3em}
\end{figure*}

\paragraph{\textbf{Architecture of LLMs.}}
We focus on the autoregressive Transformer-based LLM architecture~\citep{vaswani2017attention} based on prior works~\citep{geva-etal-2021-transformer,geva-etal-2022-transformer,elhage2021mathematical,dai-etal-2022-knowledge,meng2022locating,meng2022mass,yuksekgonul2023attention} with simplifications in certain explanations. Given an input prompt containing $T$ tokens $(t_1, \ldots, t_T)$ and each token $t_i$ belonging to a vocabulary $\mathcal{V}$, tokens are initially encoded by $d$-dimensional vectors $\mathbf{x}_i^0 \in \mathbb{R}^{d}$ using an embedding matrix $W_E \in \mathbb{R}^{|\mathcal{V}| \times d}$.

As shown in Fig.~\ref{fig:transformer}, the architecture has $L$ layers, and each layer consists of attention and MLP modules, which transform token embeddings to residual streams $(\mathbf{x}_1^\ell, \ldots, \mathbf{x}_T^\ell) \in \mathbf{X}^{\ell}$ at layer $\ell$, where $\mathbf{x}_i^\ell \in \mathbb{R}^d$. The residual stream at layer $\ell$ is a place where all attention and MLP modules at layer $\ell$ read from and write to~\citep{elhage2021mathematical}, and it is updated by the following equation for token $i$ at layer $\ell$:
\begin{equation}
\mathbf{x}_i^\ell = \mathbf{x}_i^{\ell-1} + \mathbf{a}_i^\ell + \mathbf{m}_i^\ell
\end{equation}
Here, $\mathbf{a}_i^\ell$ is the attention contribution for token $i$ and $\mathbf{m}_i^\ell$ is the MLP contribution at layer $\ell$\footnote{We omit the layer normalisation of attention and MLP modules at layer $\ell$ for simplification.}. At $L$ layer, the predicted probability distribution for the next token $\mathcal{P}(t_{T+1}|t_{1:T})$ is produced following:
\begin{equation}
    \mathcal{P}(t_{T+1}|t_{1:T})=\text{Softmax}\left(W_U\sigma (\mathbf{x}_T^L)\right)
\end{equation}
where $W_U \in  \mathbb{R}^{d\times |\mathcal{V}|}$ is unembedding matrix, $\sigma(\cdot)$ is pre-unembedding layer normalisation. 

The attention module mainly updates each token residual stream $\mathbf{x}_i^{l-1}$ by attending to all previous tokens in parallel. Specifically, the attention module contains $QK$ and $OV$ circuits, where the former operates $W_Q,W_K\in \mathbb{R}^{d\times d}$ matrices and the latter operates $W_O,W_V\in \mathbb{R}^{d\times d}$ matrices, respectively. Normally, $QK$ circuit determines the attention pattern $A^{\ell}$, i.e., where information is moved to and from the residual stream. $OV$ circuit further determines the attention output $\mathbf{a}_i^{\ell}$ based on the fixed attention pattern, i.e., what information is from the previous tokens' position to the current token position~\citep{elhage2021mathematical}:
% \vspace{-0.5em}
\begin{equation}
\mathbf{a}_{i,j}^\ell = \sum_{h=1}^H A_{i,j}^{\ell,h} (\mathbf{x}_j^{\ell-1}W_V^{\ell,h}) W_O^{\ell,h}=\sum_{h=1}^H \mathbf{r}_{i,j}^{\ell,h}
\end{equation}
where $\mathbf{a}_{i,j}^\ell$ indicates the attention contribution from token $i$ to token $j$, and $\mathbf{a}_i^\ell=\sum_{j=1}^T\mathbf{a}_{i,j}^\ell$. $\mathbf{r}_{i,j}^{l,h}$ indicates the weighted average values where token $i$ attend to token $j$ by head $h$ at the layer $\ell$, and $\mathbf{r}_i^{\ell,h}=\sum_{j=1}^T\mathbf{r}_{i,j}^{\ell,h}$ (See Appendix~\ref{app:llms_architecture} for detailed explanations about attention module). 
% For autoregressive Transformer-based LLMs, $A^{\ell,h}$ is lower triangular, representing that each token can only attend to previous tokens.

% For MLP module, it receives the $\mathbf{x}_i^{\ell-1}$ as input and updates following:
MLP module is normally treated as key-value memories~\citep{geva-etal-2021-transformer,geva-etal-2022-transformer,elhage2021mathematical,dai-etal-2022-knowledge}, where columns of $W_{\text{in}[:,i]}^\ell$ and rows of $W_{\text{out}[i,:]}^\ell$ are viewed as keys and values in Fig.~\ref{fig:transformer}, respectively. Given the input $\mathbf{x}_i^{\ell-1}$, the keys of MLP produce a vector of cofficients $\mathbf{k}_i^\ell=\gamma(W_\text{in}^\ell\mathbf{x}_i^{\ell-1})\in\mathbb{R}^{d_\text{m}}$, and they weights the corresponding values $\mathbf{v}_i^\ell$ in $W_\text{out}^\ell$ (See Appendix~\ref{app:llms_architecture} for detailed introduction about MLP):
% \vspace{-1.5mm}
\begin{equation}\label{eq:mlp_key_value}
\mathbf{m}_i^\ell=\sum_{n=1}^{d_\text{m}}\mathbf{k}_{i}^{\ell,n}\mathbf{v}_{i}^{\ell,n}
% \vspace{-1.5mm}
\end{equation}
\paragraph{\textbf{Logit lens.}}
Logit lens is a mechanistic interpretability approach to investigate the contribution of the intermediate layer representation in the autoregressive Transformer-based LLMs~\citep{logitlens2020}. Based on the architecture of LLMs above, the $\mathcal{P}(t_{T+1}|t_{1:T})$ at layer $L$ is the production of linear softmax of logits unembedded via $W_U$, which is the sum of input $\mathbf{x}_i^0$ and attention and MLP contributions at each layer $\ell$. Therefore, logit lens can be used to measure the weighted attention value of each head $\mathbf{r}_i^{\ell,h}\in\mathbb{R}^{d}$, each weighted value vector $\mathbf{k}_{i}^{\ell,n}\mathbf{v}_{i}^{\ell,n}\in\mathbb{R}^{d}$ at $n$-th dimensionality in MLP and intermediate residual stream $\mathbf{x}_i^\ell\in\mathbb{R}^d$ for token $i$:
% \vspace{-3mm}
\begin{equation}\label{eq:logit_lens}
\begin{split}
    \text{logit}_i^{\ell,h}(\mathbf{r}_i^{\ell,h})&=W_U\sigma(\mathbf{r}_i^{\ell,h})\\
\text{logit}_i^{\ell,n}(\mathbf{m}_i^{\ell,n})&=W_U\sigma(\mathbf{k}_{i}^{\ell,n}\mathbf{v}_{i}^{\ell,n}) \\
\text{logit}_i^\ell(\mathbf{x}_i^{\ell})&=W_U\sigma(\mathbf{x}_i^\ell)
\end{split}
\end{equation}

\section{Preliminaries: Zero-shot Learning with MCQs}\label{s:preliminary}
\paragraph{\textbf{Zero-shot learning.}} 
We mainly focus on the zero-shot learning regarding each GPT2 model, i.e., the input prompt is formatted as ``\textit{Question: $<$Question sample$>$ Answer Choices: $<$Multiple Choices$>$ Answer:}'', which is explained in Fig.~\ref{fig:MCQ_prompt}. After encoding the input prompt, GPT2 model will decode the next token prediction, which is expected as the correct answer choice.

\begin{table}
    
    \centering
    \resizebox{0.99\columnwidth}{!}{
    \begin{tabular}{l|c|c|c|c|c|c|c}
    \toprule[1.5pt]
      Datasets   & Train  & Test  & A (\%) & B (\%)& C (\%)& D (\%)& E (\%)\\\midrule
      IOI (2)   & - &1000 & 0 & 100 & - & - & -\\
      LD (3) & - &200  & 0 & 50 & 50 & -&- \\
      Greater (4) & - & 1000 & 0 & 33.33 & 33.33 & 33.33 & -\\
      ARC (4) & 1.12k &907 & 20.82 &  26.18 & 25.65 & 25.29 & - \\
      CSQA (5) & 9.74k &982 & 19.60 & 20.25 & 19.98 & 20.38 & 19.79 \\
    \bottomrule[1.5pt]
    \end{tabular}
    }
    \caption{The distribution of correct choices on each training dataset. IOI, LD, and Greater-than datasets are manual-synthesised and we did not choose or place the correct choice at \texttt{A}. For test datasets, we only select samples whose correct choices are not `\texttt{A}' to avoid overlap between anchored predictions from GPT2 models and the correct choice. $(\cdot)$ indicates the number of choices.}
    \label{tab:option_dist_each_dataset}
    % \vspace{-3mm}
\end{table}

\paragraph{\textbf{Datasets and models.}} 
To comprehensively verify and evaluate the anchored bias of GPT2 family, we consider 5 datasets, which include different numbers of choices from 2 to 5. \textit{Indirect Object Identification} (IOI)~\cite{wang2022interpretability} and \textit{Greater-than task} (Greater)~\cite{hanna2024does}: These two datasets have been verified that GPT2 family works well~\citep{wang2022interpretability,merullo2023circuit,hanna2024does}. 
However, we found that GPT2 family immediately fails these tasks if the input prompt is formatted as MCQ in Fig.~\ref{fig:MCQ_prompt}, where the incorrect subject of the last clause or incorrect years is placed in the `\texttt{A}' choice and the prediction is always anchored at incorrect choice `\texttt{A}'. 
\textit{Logical Deduction of the Big-Bench} (LD)\footnote{\url{https://github.com/google/BIG-bench/blob/main/bigbench/benchmark_tasks/logical_deduction/three_objects/task.json}}~\cite{srivastava2023beyond}: LD is a subtask which evaluates three-object logical deduction tasks, and it is used to measure whether model can parse information about multiple choices and their mutual relationships. \textit{ARC-Challenge} (ARC)~\cite{clark2018think} and \textit{CommensenseQA} (CSQA)~\cite{talmor2019commonsenseqa} are commonly-used MCQ benchmarks to evaluate LLMs. For each dataset, we split into 90\% \texttt{Infer.} set for anchored bias discovering and mitigation, and 10\% \texttt{Eva.} set to evaluate the modified GPT2 model performance without accessing the gold labels. For models, we comprehensively evaluate the GPT2 family, i.e., GPT2-Small-124M, GPT2-Medium-355M, GPT2-Large-774M, and GPT2-XL-1.5B~\cite{radford2019language} (See Appendix~\ref{app:datasets} for a detailed introduction of each dataset).

\paragraph{\textbf{Evaluation metrics.}} 
We use logit lens~\citep{logitlens2020} introduced in \S~\ref{s:background} to localise the specific layer of MLP and specific attention head which contribute to the anchored bias (more details in \S~\ref{s:discover}). Moreover, we use MLP contribution~\citep{geva-etal-2022-transformer} to locate the specific dimensionality from $W_\text{out}^\ell$ which leads to anchored bias. Regarding mitigating anchored bias, we use classification accuracy to evaluate whether the anchored bias can be mitigated and GPT2 family can successfully predict correct choices in MCQ.

\section{Discovering Anchored Bias in MCQs}\label{s:discover}
\paragraph{Frequency of anchored bias in GPT2 family across all datasets.}
\begin{table}[tb]
% \vspace{-7mm}
    
    \centering
    \resizebox{0.99\columnwidth}{!}{
    \begin{tabular}{l|c|c|c|c}
    \toprule[1.5pt]
      Dist. (\%)       & GPT2-Small & GPT2-Medium & GPT2-Large & GPT2-XL \\\midrule
      IOI (2)  & 45.5  & 97.4 & 100.0  & 85.8  \\
      LD (3)  & 63.0  & 94.0 & 100.0 & 17.0\\
      Greater (4)   & 32.1& 95.0& 99.5& 98.0 \\
      ARC (4)  & 54.6 & 91.6  & 97.6 &  69.9 \\
      CSQA (5) & 34.8 & 81.5& 99.6& 97.7\\
    \bottomrule[1.5pt]
    \end{tabular}
    }
    \caption{The distribution of anchored bias `\texttt{A}' happened for GPT2 family across different datasets.}
    \label{tab:dist_anchor_bias_each_dataset}
    % \vspace{-3mm}
\end{table}

As introduced in \S~\ref{s:preliminary}, we use 5 different datasets with choices from 2 to 5 to investigate the anchored bias. Table~\ref{tab:option_dist_each_dataset} shows that the correct choices distribution of ARC and CSQA training datasets is balanced from `\texttt{A}' to `\texttt{E}'. In addition, IOI, LD and Greater test datasets are manually synthesised, and we did not choose or place the correct choice at `\texttt{A}'. For all test datasets, we only select samples whose correct choice is not `\texttt{A}' to avoid introducing extra bias, i.e., the mix-up between correct prediction and anchored bias of GPT2 family. Based on the randomly sampled test datasets in Table~\ref{tab:option_dist_each_dataset}, we further calculate the distribution of anchored bias `\texttt{A}' that happened within each test dataset when different GPT2 model is used in Table~\ref{tab:dist_anchor_bias_each_dataset}. We can find that GPT2-Large and GPT2-Medium have the most serious anchored bias, and GPT2-XL and GPT2-Small have relatively less serious issues. Based on this situation, we mainly focus on investigating test samples which have anchored bias for each dataset. Table~\ref{tab:statis_test_dataset} in Appendix~\ref{app:statistic} shows the number of test samples for each dataset and GPT2 model, where \texttt{Infer.} is used to localise and mitigate anchored bias and \texttt{Eva.} is used to verify the performance of mitigation.

\begin{figure*}[tb]
    \centering
    \includegraphics[scale=0.4]{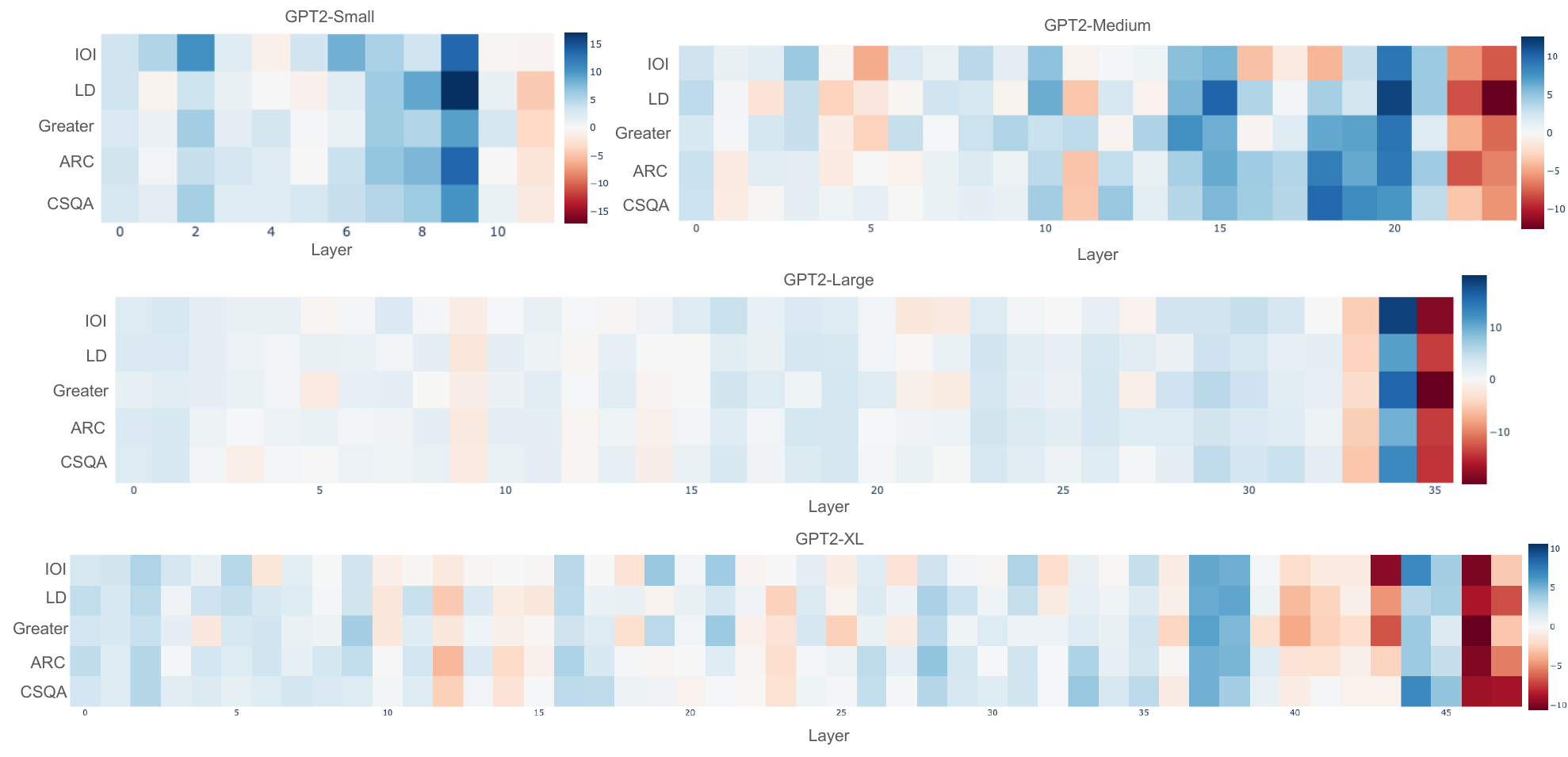}
    % \vspace{-3mm}
    \caption{MLP logit difference between anchored bias token `\texttt{A}' and correct tokens (one of $\texttt{B,C,D,E}$), i.e., $\text{logit}_T^\ell[\texttt{A}](\mathbf{m}_T^\ell) - \text{logit}_T^\ell[\texttt{B/C/D/E}](\mathbf{m}_T^\ell)$ which is averaged within GPT-2 family across all layers and all datasets. The deeper the blue blocks are at each layer, the more serious the anchored bias is, and vice versa.}
    \label{fig:gpt2_mlp_diff_lens}
    % \vspace{-1em}
\end{figure*}

\paragraph{Locating MLP of GPT2 family for anchored bias.}
We first investigate MLP modules within GPT2 family for anchored bias. Inspired from~\cite{geva-etal-2021-transformer,geva-etal-2022-transformer}, the MLP modules can be regarded as key-value memories. As introduced in \S~\ref{s:background} and Fig.~\ref{fig:transformer}, the keys of MLP module is a vector of coefficients $\mathbf{k}_i^\ell=\gamma(W_\text{in}^\ell\mathbf{x}_i^{\ell-1})\in\mathbb{R}^{d_\text{m}}$, which dynamically controls the contributions of the corresponding values $\mathbf{v}_i^\ell$ in $W_\text{out}^\ell$ based on different input prompts. The value $\mathbf{v}_i^\ell$ is treated as a memory bank which stores knowledge after the model pertaining.

Based on the consensus about MLP module, we aim to solve these research questions: 1) Is MLP responsible for the anchored bias in GPT2 family? 2) Which layer and dimensionality of MLP is anchored bias relevant to? 3) Is this bias stored as knowledge in a specific value vector of $W_\text{out}^\ell$?

We use logit lens~\citep{logitlens2020} to calculate logit of the final input prompt token contributing to incorrect choice token `\texttt{A}' and correct choice token `\texttt{B/C/D/E}' based on different datasets using Eq.~\ref{eq:mlp_key_value} and Eq.~\ref{eq:logit_lens}\footnote{The reason why we focus on the final input token is that the information inside of autoregressive transformer-based model will accumulate to the final input token for the next token prediction.}:
\begin{equation}
\begin{split}
    \text{logit}_T^\ell(\mathbf{m}_T^\ell)[\texttt{A}]&=W_U[\texttt{A}]\sigma(\mathbf{m}_T^\ell)\\
    \text{logit}_T^\ell(\mathbf{m}_T^\ell)[\texttt{B/C/D/E}]&=W_U[\texttt{B/C/D/E}]\sigma(\mathbf{m}_T^\ell)
\end{split}
\end{equation}
where $W_U[\texttt{A}]\in\mathbb{R}^{d\times|\texttt{A}|}, W_U[\texttt{B/C/D/E}]\in\mathbb{R}^{d\times|\texttt{B/C/D/E}|}$, and $|\texttt{A}|, |\texttt{B/C/D/E}|$ represents the token number index of `\texttt{A}' and one of token number index of `\texttt{B/C/D/E}', respectively.

We calculate the MLP logit difference between anchored bias token `\texttt{A}' and correct choice token \texttt{B/C/D/E} averaged across all layers and datasets for each GPT2 model using \texttt{Infer.} test samples. As shown in Fig.~\ref{fig:gpt2_mlp_diff_lens}, layer 9 in GPT2-Small, layer 20 in GPT2-Medium, layer 34 in GPT2-Large, and layer 37/38/44 in GPT2-XL are dominant layers\footnote{In this work, the layer number and head number start from 0.} related to anchored bias. In addition, these layers are much closer to the final layer of GPT2, which agrees with~\cite{geva-etal-2022-transformer,gurnee2023finding}'s finding that higher layers in GPT2 are relevant to semantic concepts or complicated tasks. We also notice that the last one or two layers in each GPT2 model do not have anchored bias at all, and they contribute more logits to the correct choice token `\texttt{B/C/D/E}' than to `\texttt{A}'. However, as the anchored bias logits are accumulated from previous layers, the final one or two layers cannot totally correct this bias.

Based on the pattern from Fig.~\ref{fig:gpt2_mlp_diff_lens}, we further use MLP contribution~\citep{geva-etal-2022-transformer} to localise the specific dimensionality from $W_\text{out}^\ell$ in these identified layers leading to anchored bias:
\begin{equation}\label{eq:mlp_contrib}
    \text{Contrib}(\mathbf{v}_T^{\ell,n})=|\mathbf{k}_T^{\ell,n}|||\mathbf{v}_T^{\ell,n}||
\end{equation}
where $|\mathbf{k}_T^{\ell,n}|$ is the absolute value of the coefficient $\mathbf{k}_T^{\ell,n}$, and $n\in d_m$. After using Eq.~\ref{eq:mlp_contrib}, we can locate the top-10 most dominant weighted value vector $\mathbf{k}_T^{\ell,n}\mathbf{v}_T^{\ell,n}$ and dimensionality with the largest contribution of the final input prompt token. We further calculate logit difference of these identified layers and dominant dimensionality in MLP of the final input token contributing anchored bias token `\texttt{A}' and correct choice tokens \texttt{B/C/D/E} using Eq.~\ref{eq:logit_lens}, i.e., $\text{logit}_T^{\ell,n}[\texttt{A}](\mathbf{m}_T^{\ell,n})-\text{logit}_T^{\ell,n}[\texttt{B/C/D/E}](\mathbf{m}_T^{\ell,n})$. Then we select candidates among the top-10 dominant dimensionality where difference score is larger than 4.
\begin{table}[tb]
    
    \centering
    \resizebox{\columnwidth}{!}{
    \begin{tabular}{l|l|l}
    \toprule[1.5pt]
         Model & Vector & Top-10 Tokens\\\midrule
        \multirow{2}{*}{GPT2-Small} & $\mathbf{v}^{9,1853}$ (100\%) & \texttt{\textvisiblespace The, \textvisiblespace This, \textvisiblespace A, \textvisiblespace There, \textvisiblespace It, \textvisiblespace In, \textvisiblespace We, \textvisiblespace If, \textvisiblespace When, \textvisiblespace An}\\
        & $\mathbf{v}^{9,2859}$ (61.8\%) & \texttt{\textvisiblespace A, \textvisiblespace In, \textvisiblespace The, \textvisiblespace (,  \textbackslash n, -, \textvisiblespace ", \textvisiblespace To, \textvisiblespace No, `} \\\midrule
        GPT2-Medium & $\mathbf{v}^{20,3713}$ (79.3\%) & \texttt{\textvisiblespace a, \textvisiblespace an, a, an, \textvisiblespace another, \textvisiblespace something, A, \textvisiblespace the, \textvisiblespace some, \textvisiblespace any}\\\midrule
        GPT2-Large & $\mathbf{v}^{34,1541}$ (100\%) & \texttt{\textvisiblespace A, A, \textvisiblespace An, \textvisiblespace Aires, \textvisiblespace Ae, \textvisiblespace An, ierrez, AAF, Aim, \textvisiblespace Aus}\\\midrule
        \multirow{2}{*}{GPT2-XL} & $\mathbf{v}^{44,4967}$ (98.0\%) & \texttt{A, \textvisiblespace A, a, AIN, aic, acebook, aa, An, AAAA, ae}\\
        & $\mathbf{v}^{38,4191}$ (100\%) & \texttt{\textvisiblespace a, \textvisiblespace an, a, \textvisiblespace of, ,, \textvisiblespace and, ., \textvisiblespace in, an, \textvisiblespace the}\\
    \bottomrule[1.5pt]
    \end{tabular}
    }
    \caption{Identified anchored-bias value vectors $\mathbf{v}^{\ell,n}$ of $n$-row of $W_\text{out}^\ell$ at layer $\ell$ for each GPT2 model, where the percentage indicates how frequently the specific $\mathbf{v}^{\ell,n}$ is detected as an anchored-bias vector across all datasets, and \texttt{\textvisiblespace} represents single space within the token because GPT2 tokeniser encodes same word with or without \texttt{\textvisiblespace} as different token numbers. For each value vector, we further unembedded the top-10 tokens, and most of them are human-interpretable words, which also verify that pretrained GPT2 family has intrinsic anchored bias within $W_\text{out}$ (See Appendix~\ref{app:top_10_tokens} for more unembedded tokens for each GPT2 model).}
    \label{tab:value_unembed}
    % \vspace{-3mm}
\end{table}
In the Table~\ref{tab:value_unembed}, the vector column demonstrates the specific value vectors $\mathbf{v}^{\ell,n}$ which are responsible for anchored bias. For each value vector, we also calculate how frequently it is recognised as an anchored-bias vector across all datasets for each GPT2 model. We can find that most identified value vectors have more than a 50\% chance with anchored bias happening across all datasets and different GPT2 models. To further verify whether these value vectors store anchored knowledge bias, we unembedded each value vector logit and selected the top-10 tokens with the highest probability. As shown in Table~\ref{tab:value_unembed}, we can find that most top-10 tokens within each value vector are relevant to `\texttt{A}', e.g., \texttt{\textvisiblespace A, A, \textvisiblespace a, a}, etc. This finding proves that some value vectors in $W_\text{out}$ of GPT2 family store knowledge bias after pertaining, and these knowledge biases will become anchored bias when the input prompt is formatted as MCQ. In addition, most unembedded tokens are stopwords, e.g., pronouns, articles, prepositions, etc, which also agrees with the findings that ``\textit{stopwords/punctuation}'' are commonly distributed in the value vectors of MLP~\citep{geva-etal-2022-transformer}.

\begin{figure*}[tb]
    \centering
    \includegraphics[scale=0.47]{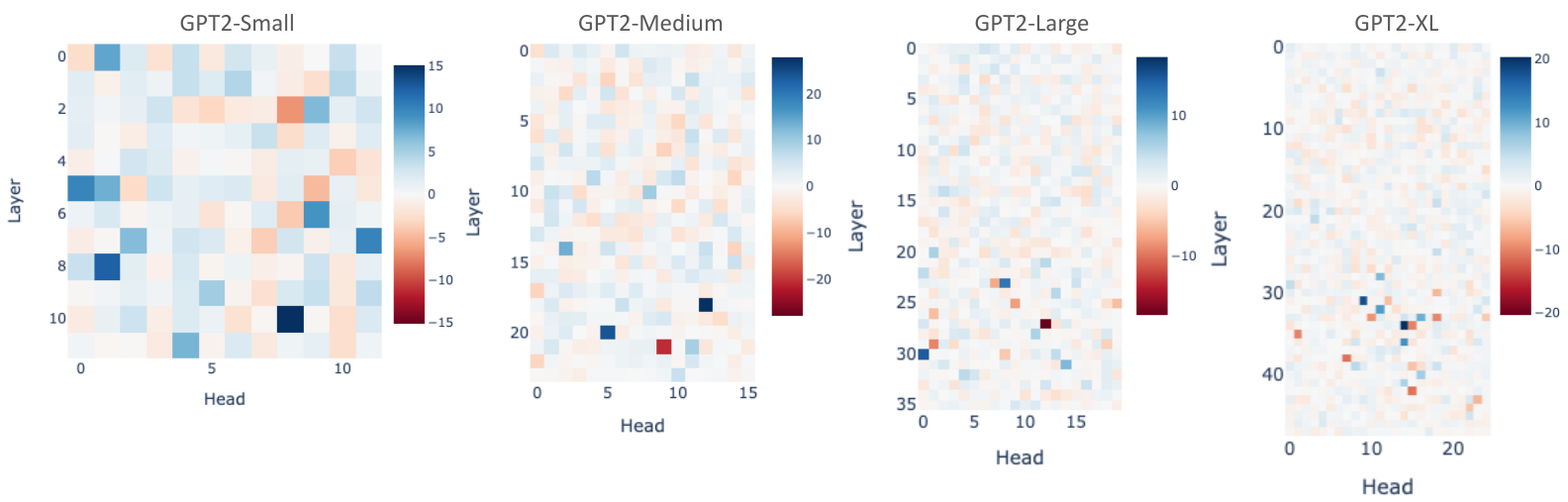}
    % \vspace{-3mm}
    \caption{Attention pattern logit difference between anchored bias token `\texttt{A}' and correct tokens (one of $\texttt{B,C,D,E}$), i.e., $\text{logit}_T^{\ell,h}[\texttt{A}](\mathbf{r}_T^{\ell,h}) - \text{logit}_T^{\ell,h}[\texttt{B/C/D/E}](\mathbf{r}_T^{\ell,h})$ which is averaged within GPT-2 family across all layers and all datasets. The deeper the blue blocks are at each layer, the more serious the anchored bias is, and vice versa.}
    \label{fig:gpt2_attn_diff_lens}
    % \vspace{-3mm}
\end{figure*}

\paragraph{Locating attention heads of GPT2 family for anchored bias.}
Following a similar method to locating anchored bias in MLP, we also aim to solve these research questions: 1) Is the attention head also responsible for the anchored bias in GPT2 family? 2) Which layer and head of attention pattern is anchored bias relevant to?

\begin{figure*}[tb]
    \centering
    \includegraphics[width=\textwidth]{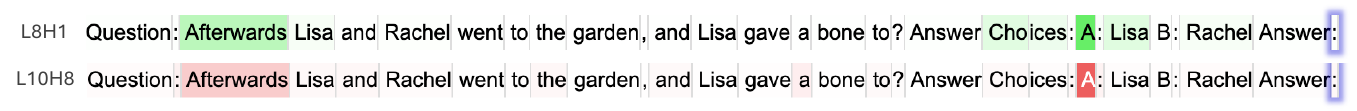}
    % \vspace{-3mm}
    \caption{The visualisation of identified anchored-bias attention head \texttt{L8H1} and \texttt{L10H8} in the GPT2-Small based on Fig.~\ref{fig:gpt2_attn_diff_lens}, where the attention weight of final token `\texttt{:}' mainly attends to `\texttt{A}' rather than `\texttt{B}'.}
    \label{fig:attn_pattern_vis_IOI}
    % \vspace{-2mm}
\end{figure*}

\begin{figure*}[tb]
    \centering
    \includegraphics[scale=0.25]{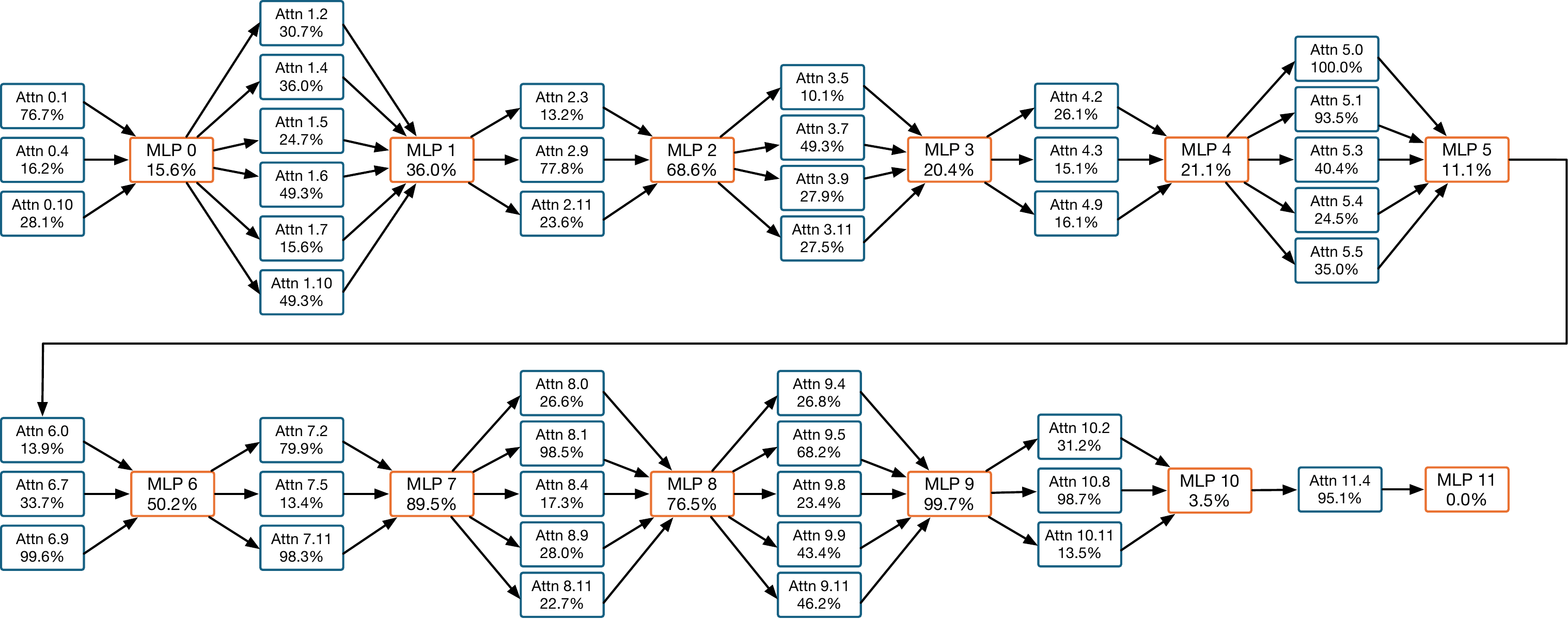}
    % \vspace{-2mm}
    \caption{The full circuit of anchored bias for GPT2-Small model, where each attention head and MLP module are selected when MLP and attention pattern logit difference threshold is larger than 4. The percentage within each module indicates the probability of anchored bias across different datasets for GPT2-Small model when threshold is larger than 4 (See Appendix~\ref{app:gpt2_curcuit} for the full circuits of other GPT2 models).}
    \label{fig:gpt2_small_circuit}
    % \vspace{-4.5mm}
\end{figure*}

\begin{table}
% \vspace{-1.5mm}
    
    \centering
    \resizebox{0.99\columnwidth}{!}{
    \begin{tabular}{l|l|l}
    \toprule[1.5pt]
         Model & Updated Vector & New Top-10 Tokens\\\midrule
        \multirow{2}{*}{GPT2-Small} & $\mathbf{v}^{9,1853}$ (100\%) & \texttt{\textvisiblespace B, B, \textvisiblespace b, \textvisiblespace C, \textvisiblespace D, \textvisiblespace P, \textvisiblespace L, \textvisiblespace R, \textvisiblespace H, \textvisiblespace F}\\
        & $\mathbf{v}^{9,2859}$ (61.8\%) & \texttt{\textvisiblespace B, B, \textvisiblespace b, \textvisiblespace C, \textvisiblespace D, \textvisiblespace L, \textvisiblespace P, \textvisiblespace R, \textvisiblespace G, \textvisiblespace F} \\\midrule
        GPT2-Medium & $\mathbf{v}^{20,3713}$ (79.3\%) & \texttt{\textvisiblespace C, C, \textvisiblespace B, \textvisiblespace c, \textvisiblespace D, \textvisiblespace G, \textvisiblespace F, \textvisiblespace P, \textvisiblespace CS, \textvisiblespace T}\\\midrule
        GPT2-Large & $\mathbf{v}^{34,1541}$ (100\%) & \texttt{\textvisiblespace C, \textvisiblespace A, \textvisiblespace B, C, \textvisiblespace c, \textvisiblespace D, \textvisiblespace F, \textvisiblespace P, \textvisiblespace G, \textvisiblespace T}\\\midrule
        \multirow{2}{*}{GPT2-XL} & $\mathbf{v}^{44,4967}$ (98.0\%) & \texttt{\textvisiblespace C, \textvisiblespace c, C, \textvisiblespace A, \textvisiblespace B, \textvisiblespace D, \textvisiblespace F, \textvisiblespace P, \textvisiblespace T, \textvisiblespace G}\\
        & $\mathbf{v}^{38,4191}$ (100\%) & \texttt{\textvisiblespace C, \textvisiblespace c, C, \textvisiblespace B, \textvisiblespace D, \textvisiblespace P, \textvisiblespace F, \textvisiblespace T, \textvisiblespace L, \textvisiblespace R}\\
    \bottomrule[1.5pt]
    \end{tabular}
    }
    \caption{The new top-10 tokens of each updated value vector for each GPT2 model (See Appendix~\ref{app:new_top_10_tokens} for more new unembedded tokens for each GPT2 model).}
    \label{tab:new_value_unembed}
    % \vspace{-4mm}
\end{table}
As explained in \S~\ref{s:background}, attention pattern $\mathbf{r}_{i,j}^{\ell,h}$ indicates the weighted average values where token $i$ attend to token $j$ by head $h$ at the layer $\ell$. We use logit lens to analyse the logit difference of final input prompt token contribution between anchored bias `\texttt{A}' and correct choices `\texttt{B/C/D/E}', i.e., $\text{logit}_T^{\ell,h}[\texttt{A}](\mathbf{r}_T^{\ell,h}) - \text{logit}_T^{\ell,h}[\texttt{B/C/D/E}](\mathbf{r}_T^{\ell,h})$. As shown in Fig.~\ref{fig:gpt2_attn_diff_lens}, L8H1 and L10H8 in GPT2-Small, L18H12 and L20H5 in GPT2-Medium, L23H8 and L30H0 in GPT2-Large, L31H9 and L34H14\footnote{L34H14 indicates layer 34 and head 14 and both of them start from 0.} in GPT2-XL are dominant heads related to anchored bias. Those heads are also distributed closer to the final layer in each GPT2 model. We zoom in on the L8H1 and L10H8 attention pattern of the final input token in GPT2-Small using a sample from IOI dataset. As shown in Fig.~\ref{fig:attn_pattern_vis_IOI}, the final token `\texttt{:}' attends more weights on the anchored bias token `\texttt{A}' than the correct choice token `\texttt{B}', which agrees with our identified attention head using logit difference in Fig.~\ref{fig:gpt2_attn_diff_lens}. In addition, the full circuit of anchored bias for each GPT2 model can be built based on the MLP and attention logit difference in Fig.~\ref{fig:gpt2_small_circuit}.

\section{Mitigating Anchored Bias in MCQs}\label{sec:mitigate_bias}
\paragraph{Mitigating anchored bias in MLP.}

According to findings in \S~\ref{s:discover}, we localise the specific value vector in MLP related to the anchored bias. We further aim to solve the following research question: Can we fix the identified value vector in MLP by updating its values and editing the biased knowledge?

Following~\citep{dai-etal-2022-knowledge}, we directly modify and update the identified value vector as\footnote{The $[\texttt{A}]$ and $[\texttt{B/C/D/E}]$ represent the corresponding token index number from the tokenizer vocabulary. Then $W_U[\texttt{B/C/D/E}]$ means that we extract the token unembedding vector from $W_U$ for token \texttt{B} or \texttt{C} or \texttt{D} or \texttt{E}. For example, if we assume the token index number of `\texttt{B}' is 358 in the vocabulary, we will extract the 358-th column from $W_U\in\mathbb{R}^{d \times| \mathcal{V}|}$. Then we use Eq.~\ref{eq:mlp_update} to update the value vector by adding back the unembedding vector multiplied by 
 to the value vector, where the shape of the value vector and the unembedding vector are the same.}:
\begin{equation}\label{eq:mlp_update}
    \mathbf{v}^{\ell,n}=\mathbf{v}^{\ell,n} - \lambda_1 W_U[\texttt{A}] + \lambda_2 W_U[\texttt{B/C/D/E}]
\end{equation}
where $\lambda_1=1, \lambda_2=8$. After updating the corresponding value vector in MLP, we utilise the updated GPT2 model to predict the next token of the same input MCQ prompts. We comprehensively evaluate each updated GPT2 model with the corresponding modified value vector using \texttt{Infer.} and \texttt{Eva.} across all datasets. As shown in Table~\ref{tab:fix_mlp_success_rate}, most updated value vectors achieve high classification accuracy regarding MCQ tasks, even with multiple near 100\% or 100\% accuracy in $\mathbf{v}^{9,1853}$ of GPT2-Small, $\mathbf{v}^{34,1541}$ of GPT2-Large, $\mathbf{v}^{44,4967}$ and $\mathbf{v}^{38,4191}$ of GPT2-XL. For those value vectors with around a 60-70\% chance of anchored bias happening across all datasets and different GPT2 models (i.e., $\mathbf{v}^{9,2859}$ of GPT2-Small, $\mathbf{v}^{20,3713}$ of GPT2-Medium, $\mathbf{v}^{34,2103}$ of GPT2-Large, and $\mathbf{v}^{37,2966}$ of GPT2-XL in Appendix~\ref{app:mlp_ablate}), the classification accuracy still 68.09\% averaged all datasets and all models. This means that the simple and straightforward method (i.e., Eq.~\ref{eq:mlp_update}) is effective, and we do not need to fine-tune the whole GPT2 model to fix the anchored bias. In addition, such performance gains can be generalised to the \texttt{Eva.} dataset without accessing gold labels. We further unembedded the updated value vectors in Table~\ref{tab:new_value_unembed}, and it shows that the anchored bias token `\texttt{A}' is significantly removed and the new top-10 tokens for each new value vector are replaced with the correct choices token `\texttt{B/C/D/E}'.

\begin{table}[tb]
    
    \centering
    \resizebox{0.99\columnwidth}{!}{
    \begin{tabular}{l|l|cc|cc|cc|cc|cc}
    \toprule[1.5pt]
         \multirow{2}{*}{Model} & \multirow{2}{*}{Vector} & \multicolumn{2}{c|}{IOI (2)} & \multicolumn{2}{c|}{LD (3)} & \multicolumn{2}{c|}{Greater (4)} & \multicolumn{2}{c|}{ARC (4)} & \multicolumn{2}{c}{CSQA (5)}\\
         & & Infer. & Eva. & Infer. & Eva. & Infer. & Eva. & Infer. & Eva.& Infer. & Eva.\\\midrule
        \multirow{2}{*}{GPT2-Small} & $\mathbf{v}^{9,1853}$ (100\%) & 100.0 & 100.0 & 100.0 & 100.0 & 100.0 & 100.0& 100.0& 100.0 &100.0 &100.0\\
        & $\mathbf{v}^{9,2859}$ (61.8\%) & 44.6 & 44.2 & 100.0 &100.0 & 58.8 & 92.3 &93.7 & 96.2 & 60.1& 56.4\\\midrule
        GPT2-Medium & $\mathbf{v}^{20,3713}$ (79.3\%) & 99.4 & 98.1 &97.1 & 88.5 &30.6 & 70.2 &56.8 &60.4 &26.9 & 27.5\\\midrule
        GPT2-Large & $\mathbf{v}^{34,1541}$ (100\%) & 100.0 &100.0 &100.0 &100.0 &100.0 & 100.0&96.7 &96.7 & 99.7&100.0\\\midrule
        \multirow{2}{*}{GPT2-XL} & $\mathbf{v}^{44,4967}$ (98.0\%) & 98.2 & 97.8&100.0 & 100.0&100.0 &100.0 &90.7 &90.8 &94.8 &94.2\\
        & $\mathbf{v}^{38,4191}$ (100\%) & 100.0 &100.0 &100.0 & 100.0&94.9 &100.0 &97.4 &96.9 &96.5 &95.2\\
    \bottomrule[1.5pt]
    \end{tabular}
    }
    \caption{The classification accuracy of each MCQ inference (Infer.) and evaluation (Eva.) dataset after updating the identified value vectors for each GPT2 model using $\lambda_2=8$ (See Appendix~\ref{app:mlp_ablate} for the ablation study of $\lambda_2$ with different values).}
    \label{tab:fix_mlp_success_rate}
    % \vspace{-1.5mm}
\end{table}

\paragraph{Mitigating anchored bias in attention heads.}
Based on the located attention heads for each GPT2 model in \S~\ref{s:discover}, we follow the same pattern as fixing anchored bias in MLP and further propose a recalibration approach to mitigate the anchored bias in the attention head by swapping the attention weight of $\mathbf{r}_T^{\ell,h}$ between the position of `\texttt{A}' and `\texttt{B/C/D/E}':
\begin{equation}
    \mathbf{r}_{T,p(\texttt{A})}^{\ell,h} = \mathbf{r}_{T,p(\texttt{B/C/D/E})}^{\ell,h}\quad \mathbf{r}_{T,p(\texttt{B/C/D/E})}^{\ell,h}=\mathbf{r}_{T,p(\texttt{A})}^{\ell,h}
\end{equation}
where $p(\texttt{A})$ and $p(\texttt{B/C/D/E})$ indicate the actual position of anchored bias token `\texttt{A}' and correct choices token `\texttt{B/C/D/E}' in the input prompts. As shown in Table~\ref{tab:mitigate_attn_acc}, the attention recalibration works in L18H12 of GPT2-Medium, especially for IOI dataset. This finding means that MLP module plays an important role than the attention head for the anchored bias, and the performance of attention recalibration depends on the choice of GPT2 model and dataset.

\begin{table}[tb]
    
    \centering
    \resizebox{0.99\columnwidth}{!}{
    \begin{tabular}{l|l|cccccccccc}
    \toprule[1.5pt]
         \multirow{2}{*}{Model} & \multirow{2}{*}{Head} & \multicolumn{2}{c}{IOI (2)} & \multicolumn{2}{c}{LD (3)} & \multicolumn{2}{c}{Greater (4)} & \multicolumn{2}{c}{ARC (4)} & \multicolumn{2}{c}{CSQA (5)}\\
         & & Infer. & Eva. & Infer. & Eva. & Infer. & Eva. & Infer. & Eva.& Infer. & Eva.\\\midrule
        GPT2-Medium & L18H12 & 92.47 & 90.72 & 21.76 & 27.78 & 3.39 & 3.16 & 24.33 & 21.69 & 1.67 & 1.25\\\midrule
        \multirow{2}{*}{GPT2-XL} & L31H9 & 0.0 & 0.0 & 9.68 & 0.0 & 0.57 & 0.0 & 0.53 & 0.0 & 0.12 & 0.0\\
        & L34H14 & 0.0 & 0.0 & 12.90 & 0.0 & 0.57 & 0.0 &  3.33& 1.67 & 0.46 & 0.0\\
    \bottomrule[1.5pt]
    \end{tabular}
    }
    \caption{The Top-1 classification accuracy changes after attention pattern recalibration for each GPT2 model across all datasets (See Appendix~\ref{app:attn_recali} for other GPT2 model's results).}
    \label{tab:mitigate_attn_acc}
    % \vspace{-4.5mm}
\end{table}

% \vspace{-0.5em}
\section{Discussion}\label{s:discussion}
\paragraph{\textbf{Is Few-shot learning helpful?}}
Based on the comprehensive zero-shot learning MCQ across all datasets, we have a good understanding of how important the MLP and attention head are regarding the anchored bias in the GPT2 family. The following question might be whether few-shot learning can mitigate this anchored bias without updating specific value vectors in MLP or recalibrating the attention head. We conduct an experiment to evaluate GPT2 family across all datasets using 1-shot and 2-shot learning settings. The initial finding is that the anchored bias could be relatively mitigated and average MCQ classification accuracy across GPT2 family is 46.74\%, 44.44\%, 38.46\% and 23.34\% under 1-shot learning and 46.52\%, 45.70\%, 43.22\% and 32.62\% under 2-shot learning (See Appendix~\ref{app:few_shot}). This result indicates that GPT2 family still struggles to predict correct choices, especially for GPT2-XL, which needs more investigation in the future.
% to a certain extent depending on the length of few-shot prompts and the complexity of the few-shot prompt content. However, the anchored bias still happens, especially for the long and complex few-shot prompts. Even if the anchored bias to `\texttt{A}' disappears, GPT2 model cannot still predict the correct choice and generate other incorrect choices, which needs more investigation. We leave this question as the future work.

\begin{table}[tb]
    \centering
    \resizebox{0.99\columnwidth}{!}{
    \begin{tabular}{l|ccc}
    \toprule
       Models  & CNN (Rouge-Avg) & WikiText2 (PPL) & PTB (PPL) \\\hline
        GPT2-Small & 16.8 / 15.2 & 31.2 / 33.5 & 66.1 / 67.4 \\
        GPT2-Medium & 20.4 / 18.6 & 23.8 / 25.9 & 48.2 / 49.7 \\
        GPT2-Large & 21.5 / 19.9 & 20.5 / 22.3 & 41.8 / 43.2 \\
        GPT2-XL & 21.7 / 20.2 & 19.1 / 20.1 & 36.2 / 37.9 \\
    \bottomrule
    \end{tabular}}
    \caption{The performance of different GPT2 models on three different tasks with random-selected 1000 samples before and after updating the identified MLP value vectors in the~\S~\ref{sec:mitigate_bias}. We use / to separate the performance before and after the intervention.}
    \label{tab:harm_other_tasks}
\end{table}

\paragraph{\textbf{Is direct value vector updating in MLP harmful to the general ability of GPT2 for other tasks?}}
 We conduct an experiment to evaluate whether direct value vector updating in MLP is harmful to the general ability of GPT2-Small for the original IOI and Greater-than tasks. The experiment shows that the modified GPT2 family can still achieve the average 85.8\%, 91.1\%, 81.7\% and 78.8\% accuracy on the original IOI dataset, and 96.1\%, 98.0\%, 98.5\% and 98.4\% on the original Greater-than dataset (See Appendix~\ref{app:damage}). Although direct value vector updating in MLP is harmful to the general ability of GPT2 on the original IOI and Greater-than datasets, this model editing approach does not produce serious damage, which matches the findings from~\citet{gu-etal-2024-model}. Similarly, we further evaluate the general ability of GPT-2 families on the CNN and Daily Mail, WikiText2 and PTB tasks using the identified value vector of MLP in the~\S~\ref{sec:mitigate_bias}. Table~\ref{tab:harm_other_tasks} demonstrates similar findings to the Appendix~\ref{app:damage}. However, we need to develop a better and minimal-harm model editing algorithm in the future.

 \paragraph{Is anchored bias sensitive to specific content of input prompts?}
 We construct two random MCQ datasets which include random concatenated characters and random vocabulary words (see Table~\ref{tab:random_MCQ_prompt} and Table~\ref{tab:random_MCQ_prompt_example} in Appendix~\ref{app:random_words}). The result shows that anchored bias still happens across different GPT2 models, especially for GPT2-Large and GPT2-XL. This indicates that anchored bias is insensitive to MCQ input prompts, which also confirms our findings that GPT2 family exhibit the anchored bias with significant regularity across various MCQ datasets.

\section{Conclusion}
In this work, we identify the anchored bias of GPT2 family, where GPT-2 models consistently favour the first choice `\texttt{A}' in the MCQ task. Based on this observation, we comprehensively conduct a mechanistic analysis of the internal workings of GPT2 family. We find that some value vectors in MLP modules with specific layers and dimensionality play a significant role in the anchored bias, and we further use a straightforward but potent approach to update the corresponding value vectors, which effectively mitigate the anchored bias in GPT2 family. In addition, some attention heads also play auxiliary roles in this bias, and the recalibration approach works well for the IOI dataset in GPT2-Medium.
\section*{Acknowledgement}
We would like to thank all the anonymous reviewers for their insightful comments. This work is supported by the Gemma 2 Academic Program GCP Credit Award from Google.
\section*{Limitations}\label{app:limitation}
This work mainly focuses on the mechanistic analysis of GPT2 family with the model size from 124M to 1.5B. It is worth comprehensively investigating whether larger open-source LLMs have similar anchored biases, such as LLaMA-7B-65B, LLaMA2-7B-70B, LLaMA3-8B-71B, etc. In addition, different LLM architectural backbones might have different extents of anchored bias, e.g., Mixture of Experts (MoE) and Mamba with selective state spaces. It is meaningful to compare how different MLPs and attention heads are across different LLMs above and why anchored bias disappears if larger LLMs do not have such an issue. Moreover, the knowledge editing approach by directly updating value vectors from MLPs is not optimal as it will introduce some extent of damage to the general ability of GPT2 models. However, how to develop a better and minimal-harm model editing algorithm is an open question~\citep{gu-etal-2024-model}, which is worth exploring in the future.
% Bibliography entries for the entire Anthology, followed by custom entries
%\bibliography{anthology,custom}
% Custom bibliography entries only
\bibliography{custom}

\appendix

\section{Broader Impacts}\label{app:impact}
Mechanistic interpretability of anchored bias for the GPT2 family under the MCQ setting is worth investigating, as it can help us better understand the inner working mechanism of MLPs and attention heads for autoregressive Transformer-based LLMs. The identified MLPs and attention heads leading to the anchored bias can be used to guide the larger LLMs development for safer, less biased and more trustworthy LLMs. Such a mechanistic analysis approach can be extended to other tasks, such as LLMs mathematical reasoning, dialogue generation, and different training methods, such as chain-of-thoughts (CoTs), reinforcement learning from human feedback (RLHF), direct preference optimization. In addition, an adversarial attack might be used for commercial LLM products when this anchored bias is analysed. This also encourages researchers to develop much safer and robust LLMs.

\section{Detailed Explanations of LLMs Architecture}\label{app:llms_architecture}

We focus on the autoregressive Transformer-based LLM architecture~\citep{vaswani2017attention} based on prior works~\citep{geva-etal-2021-transformer,geva-etal-2022-transformer,elhage2021mathematical,dai-etal-2022-knowledge,meng2022locating,meng2022mass,yuksekgonul2023attention} with simplifications in certain explanations. Given an input prompt containing $T$ tokens $(t_1, \ldots, t_T)$ and each token $t_i$ belonging to a vocabulary $\mathcal{V}$, tokens are initially encoded by $d$-dimensional vectors $\mathbf{x}_i^0 \in \mathbb{R}^{d}$ using an embedding matrix $W_E \in \mathbb{R}^{|\mathcal{V}| \times d}$.

The architecture has $L$ layers, and each layer consists of attention and MLP modules, which transform token embeddings to residual streams $(\mathbf{x}_1^\ell, \ldots, \mathbf{x}_T^\ell) \in \mathbf{X}^{\ell}$ at layer $\ell$, where $\mathbf{x}_i^\ell \in \mathbb{R}^d$. The residual stream at layer $\ell$ is a place where all attention and MLP modules at layer $\ell$ read from and write to~\citep{elhage2021mathematical}, and it is updated by the following equation for token $i$ at layer $\ell$:
\begin{equation}
\mathbf{x}_i^\ell = \mathbf{x}_i^{\ell-1} + \mathbf{a}_i^\ell + \mathbf{m}_i^\ell
\end{equation}
Here, $\mathbf{a}_i^\ell$ is the attention contribution for token $i$ and $\mathbf{m}_i^\ell$ is the MLP contribution at layer $\ell$\footnote{We omit the layer normalisation of attention and MLP modules at layer $\ell$ for simplification.}. At $L$ layer, the predicted probability distribution for the next token $\mathcal{P}(t_{T+1}|t_{1:T})$ is produced following:
\begin{equation}
    \mathcal{P}(t_{T+1}|t_{1:T})=\text{Softmax}\left(W_U\sigma (\mathbf{x}_T^L)\right)
\end{equation}
where $W_U \in  \mathbb{R}^{d\times |\mathcal{V}|}$ is unembedding matrix, $\sigma(\cdot)$ is pre-unembedding layer normalisation. 

The attention module mainly updates each token residual stream $\mathbf{x}_i^{l-1}$ by attending to all previous tokens in parallel. Specifically, the attention module contains $QK$ and $OV$ circuits, where the former operates $W_Q,W_K\in \mathbb{R}^{d\times d}$ matrices and the latter operates $W_O,W_V\in \mathbb{R}^{d\times d}$ matrices, respectively. Normally, $QK$ circuit determines the attention pattern $A^{\ell}$, i.e., where information is moved to and from the residual stream. $OV$ circuit further determines the attention output $\mathbf{a}_i^{\ell}$ based on the fixed attention pattern, i.e., what information is from the previous tokens' position to the current token position~\citep{elhage2021mathematical}:
\begin{equation}
A^{\ell,h} = \text{Softmax}\left(\frac{(\mathbf{X}^{\ell-1}W_Q^{\ell,h})(\mathbf{X}^{\ell-1}W_K^{\ell,h})^T}{\sqrt{d_h}}\right)
\end{equation}
\begin{equation}
\mathbf{a}_{i,j}^\ell = \sum_{h=1}^H A_{i,j}^{\ell,h} (\mathbf{x}_j^{\ell-1}W_V^{\ell,h}) W_O^{\ell,h}=\sum_{h=1}^H \mathbf{r}_{i,j}^{\ell,h}
\end{equation}
where $\mathbf{a}_{i,j}^\ell$ indicates the attention contribution from token $i$ to token $j$, and $\mathbf{a}_i^\ell=\sum_{j=1}^T\mathbf{a}_{i,j}^\ell$. Attention pattern $A^{\ell,h} \in \mathbb{R}^{T \times T}$ is a lower triangular weight matrix calculated by the $h$-th attention head at layer $\ell$, representing that each token can only attend to previous tokens within autoregressive Transformer-based LLMs. All matrices are split into multiple attention heads, i.e., $W_Q^{\ell,h},W_K^{\ell,h},W_V^{\ell,h}\in \mathbb{R}^{d\times d_h}$, and $W_O^{\ell,h}\in \mathbb{R}^{d_h\times d}$ for head $h$. $d_h$ is the dimensionality of each head, $H$ represents the total number of attention heads, and $d_h=d/H$. $A_{i,j}^{\ell,h}$ is the $i$-th row and $j$-th column entry of $A^{\ell,h}$, and $\mathbf{r}_{i,j}^{l,h}$ indicates the weighted average values where token $i$ attend to token $j$ by head $h$ at the layer $\ell$, and $\mathbf{r}_i^{\ell,h}=\sum_{j=1}^T\mathbf{r}_{i,j}^{\ell,h}$. 
% For autoregressive Transformer-based LLMs, $A^{\ell,h}$ is lower triangular, representing that each token can only attend to previous tokens.

For MLP module, it receives the $\mathbf{x}_i^{\ell-1}$ as input and updates following:
\begin{equation}\label{eq:mlp}       
 \mathbf{m}_i^\ell=\gamma(W_\text{in}^\ell\mathbf{x}_i^{\ell-1})W_\text{out}^\ell
\end{equation}
where $\gamma(\cdot)$ is activation function, $W_\text{in}^\ell\in\mathbb{R}^{d\times d_\text{m}}$, and $W_\text{out}^\ell\in\mathbb{R}^{d_\text{m}\times d}$. $d_\text{m}$ is the dimensionality of MLP module, which is larger than $d$. MLP module is normally treated as key-value memories~\citep{geva-etal-2021-transformer,geva-etal-2022-transformer,elhage2021mathematical,dai-etal-2022-knowledge}, where columns of $W_{\text{in}[:,i]}^\ell$ and rows of $W_{\text{out}[i,:]}^\ell$ are viewed as keys and values, respectively. Given the input $\mathbf{x}_i^{\ell-1}$, the keys of MLP produce a vector of cofficients $\mathbf{k}_i^\ell=\gamma(W_\text{in}^\ell\mathbf{x}_i^{\ell-1})\in\mathbb{R}^{d_\text{m}}$, and they weights the corresponding values $\mathbf{v}_i^\ell$ in $W_\text{out}^\ell$. Therefore, Eq.~\ref{eq:mlp} can be reformatted as\footnote{We omit all bias $b_Q,b_K,b_O,b_V,b_\text{in},b_\text{out},b_\text{U}$ for simplification.}:
\begin{equation}\label{eq:mlp_key_value_app}
\mathbf{m}_i^\ell=\sum_{n=1}^{d_\text{m}}\mathbf{k}_{i}^{\ell,n}\mathbf{v}_{i}^{\ell,n}
\end{equation}

\section{Details of Each Dataset and Experimental Settings}\label{app:datasets}
\begin{itemize}
    \item \textit{Indirect Object Identification} (IOI)~\cite{wang2022interpretability}: IOI is a manually synthesised corpus used to understand a specific natural language task, where sentences such as ``Afterwards Lisa and Rachel went to the garden, and Lisa gave a bone to'' should be followed with ``Rachel''. When two names are included in such sentences, the predicted name should not be the subject of the last clause. This task has been verified that GPT2 family works well~\citep{wang2022interpretability,merullo2023circuit}. However, we found that GPT2 family immediately fails this task if the input prompt is formatted as MCQ, where the incorrect subject of the last clause is placed in the `\texttt{A}' choice.
    \item \textit{Greater-than task} (Greater)~\cite{hanna2024does}: Greater-than task is also a manually synthesised corpus, which is used to evaluate GPT2's ability to sentences such as ``The war lasted from the year 1732 to the year 17'', and model will predict valid two-digit end years, i.e., years > 32. However, we found the same anchored bias like IOI when this task is formatted as MCQ in Fig.~\ref{fig:MCQ_prompt}, and GPT2 family also fails to predict valid years and the prediction is always anchored at incorrect choice `\texttt{A}'. 
    \item \textit{Logical Deduction of the Big-Bench} (LD)\footnote{\url{https://github.com/google/BIG-bench/blob/main/bigbench/benchmark_tasks/logical_deduction/three_objects/task.json}}~\cite{srivastava2023beyond}: LD is a subtask which evaluates three-object logical deduction tasks, and it is used to measure whether model can parse information about multiple choices and their mutual relationships. Each MCQ in the task includes three similar objects in a naturally ordered context (e.g., books of various colours sitting on a shelf) and a set of simple clues regarding their placement (e.g., "the red book is to the right of the green book") such that no clue is redundant. The challenge is to assign the highest probability to correct MCQ choice about which object lies at which position.
    \item \textit{ARC-Challenge} (ARC)~\cite{clark2018think}: ARC is a real grade-school level, multiple-choice science questions, which includes Easy and Challenge versions. We choose the Challenge version to evaluate the anchored bias.
    \item \textit{CommensenseQA} (CSQA)~\cite{talmor2019commonsenseqa}: CSQA is a  new multiple-choice question answering dataset that requires different types of commonsense knowledge to predict the correct answers.
\end{itemize}
All GPT2 models are run during inference time and parameters inside of each GPT2 model are frozen. All experiments can be easily run using CPU or GPU, e.g., Apple Macbook Pro with M1 Pro chip or NVIDIA 3090 Ti with 24GB GPU RAM.

\section{The Statistic Information of Each Test Dataset for GPT2 Models}\label{app:statistic}
We split each dataset into 90\% \texttt{Infer.} set for anchored bias discovering and mitigation, and 10\% \texttt{Eva.} set for modified GPT2 model verification on the MCQ task. Table~\ref{tab:statis_test_dataset} shows the number of test data samples for each dataset.

\begin{table}[ht]
    
    \centering
    \resizebox{\columnwidth}{!}{
    \begin{tabular}{l|cc|cc|cc|cc}
    \toprule[1.5pt]
      \multirow{2}{*}{Num.}   & \multicolumn{2}{c|}{GPT2-Small}  & \multicolumn{2}{c|}{GPT2-Medium}  & \multicolumn{2}{c|}{GPT2-Large} & \multicolumn{2}{c}{GPT2-XL}\\
      & Infer. & Eva. & Infer. & Eva. & Infer. & Eva. & Infer. & Eva.\\\midrule
      IOI (2)   & 410 & 45 & 877 & 97 & 900 & 100 & 773 & 85\\
      LD (3) & 114 & 12  & 170 & 18 & 180 & 20& 31 & 3\\
      Greater (4) & 289 & 32 & 856 & 95 & 897 & 99 & 883 & 98\\
      ARC (4) & 446 & 49 & 748 &  83 & 797 & 88 & 571 & 63\\
      CSQA (5) & 308 & 34 & 720 & 80 & 881 & 97 & 864 & 95\\
    \bottomrule[1.5pt]
    \end{tabular}
    }
    \caption{Statistic of the test datasets for each GPT2 model, where \texttt{Infer.} represents 90\% test dataset used to discover and mitigate anchored bias, and \texttt{Eva.} represents 10\% test dataset used to verify the performance of updated GPT2 models.}
    \label{tab:statis_test_dataset}
\end{table}

\section{Top-10 Unembeded Tokens from Identified Value Vectors}\label{app:top_10_tokens}
\begin{table*}[ht]
    
    \centering
    \resizebox{0.99\textwidth}{!}{
    \begin{tabular}{l|l|l}
    \toprule[1.5pt]
         Model & Vector & Top-10 Tokens\\\midrule
        GPT2-Small& $\mathbf{v}^{9,788}$ (50.0\%)& \texttt{\textvisiblespace and, \textvisiblespace in, \textvisiblespace to, \textvisiblespace or, \textvisiblespace a, \textvisiblespace at, \textvisiblespace the, \textvisiblespace ..., \textvisiblespace that, "} \\\midrule
        GPT2-Medium& $\mathbf{v}^{20,1731} (38.2\%)$ & \texttt{\textvisiblespace first, \textvisiblespace First, first, \textvisiblespace FIRST, First, \textvisiblespace 1, 1, \textvisiblespace Firstly, Firstly, \textvisiblespace begin}\\\midrule
        \multirow{2}{*}{GPT2-Large}& $\mathbf{v}^{34,2103} (70.1\%)$ & \texttt{romeda, maxwell, \begin{CJK}{UTF8}{goth}ドラゴン\end{CJK}, lvl, mobi, elaide, amsung, nil, 911, dylib}\\
        & $\mathbf{v}^{34,4178}$ (55.4\%) & \texttt{reply, interstitial, \textvisiblespace emer, 76561, \textvisiblespace err, sg, \textvisiblespace whe, eers, oi, \textvisiblespace ignor}\\\midrule
        \multirow{5}{*}{GPT2-XL}& $\mathbf{v}^{44,2995}$ (27.5\%) & \texttt{ggles, atchewan, \textvisiblespace aw, \textvisiblespace let, eed, \textvisiblespace ont, wn, \textvisiblespace be, gg, \textvisiblespace palp}\\
        & $\mathbf{v}^{44,128}$ (22.7\%) & \texttt{\textvisiblespace A, \textvisiblespace C, \textvisiblespace E, \textvisiblespace B, \textvisiblespace S, \textvisiblespace G, \textvisiblespace F, \textvisiblespace P, \textvisiblespace D, \textvisiblespace K}\\
        & $\mathbf{v}^{38,4174}$ (20.2\%) & \texttt{\textvisiblespace A, ,, \textvisiblespace and, \textvisiblespace a, \textvisiblespace at, ., \textvisiblespace of, \textvisiblespace as, \textvisiblespace on, \textvisiblespace (}\\
        & $\mathbf{v}^{37,423}$ (91.8\%) & \texttt{\textvisiblespace The, \textvisiblespace This, \textvisiblespace It, \textvisiblespace There, \textvisiblespace A, \textvisiblespace If, \textvisiblespace You, \textvisiblespace We, \textvisiblespace These, \textvisiblespace When}\\
        & $\mathbf{v}^{37,2966}$ (66.2\%) & \texttt{\textvisiblespace a, \textvisiblespace an, ,, \textvisiblespace and, \textvisiblespace to, \textvisiblespace the, \textvisiblespace in, \textvisiblespace (, \textvisiblespace one, .}\\
    \bottomrule[1.5pt]
    \end{tabular}
    }
    \caption{Identified anchored-bias value vectors $\mathbf{v}^{\ell,n}$ of $n$-row of $W_\text{out}^\ell$ at layer $\ell$ for each GPT2 model, where the percentage indicates how frequently the specific $\mathbf{v}^{\ell,n}$ is detected as an anchored-bias vector across all datasets, and \texttt{\textvisiblespace} represents single space within the token because GPT2 tokeniser encodes same word with or without \texttt{\textvisiblespace} as different token numbers. For each value vector, we further unembeded the top-10 tokens, and most of them are human-interpretable words, which also verify that pretrained GPT2 family has intrinsic anchored bias within $W_\text{out}$.}
    \label{tab:value_unembed_more}
\end{table*}
We unembedded more identified anchored-bias value vectors in Table~\ref{tab:value_unembed_more}, and we can find there are several tokens related to \texttt{`A'}, such as \texttt{\textvisiblespace a, \textvisiblespace first, \textvisiblespace First, \textvisiblespace 1, \textvisiblespace A}.

\section{Top-10 Unembedded Tokens from Updated Value Vectors}\label{app:new_top_10_tokens}
\begin{table}[ht]
    
    \centering
    \resizebox{\columnwidth}{!}{
    \begin{tabular}{l|l|l}
    \toprule[1.5pt]
         Model & Updated Vector & New Top-10 Tokens\\\midrule
        GPT2-Small& $\mathbf{v}^{9,788}$ (50.0\%)& \texttt{\textvisiblespace B, \textvisiblespace b, B, \textvisiblespace C, \textvisiblespace D, \textvisiblespace L, \textvisiblespace P, \textvisiblespace R, \textvisiblespace G, \textvisiblespace BC} \\\midrule
        GPT2-Medium& $\mathbf{v}^{20,1731}$ (38.2\%) & \texttt{\textvisiblespace C, C, \textvisiblespace B, \textvisiblespace c, \textvisiblespace D, \textvisiblespace CS, \textvisiblespace F, \textvisiblespace P, \textvisiblespace G, \textvisiblespace T}\\\midrule
        \multirow{2}{*}{GPT2-Large}& $\mathbf{v}^{34,2103}$ (70.1\%) & \texttt{\textvisiblespace C, \textvisiblespace c, C, \textvisiblespace B, \textvisiblespace D, \textvisiblespace F, \textvisiblespace P, \textvisiblespace G, \textvisiblespace T, \textvisiblespace M}\\
        & $\mathbf{v}^{34,4178}$ (55.4\%) & \texttt{\textvisiblespace C, \textvisiblespace c, C, \textvisiblespace B, \textvisiblespace P, \textvisiblespace D, \textvisiblespace F, \textvisiblespace G, \textvisiblespace T, \textvisiblespace L}\\\midrule
        \multirow{5}{*}{GPT2-XL}& $\mathbf{v}^{44,2995}$ (27.5\%) & \texttt{\textvisiblespace C, \textvisiblespace c, C, \textvisiblespace B, \textvisiblespace D, \textvisiblespace P, \textvisiblespace T, \textvisiblespace F, \textvisiblespace R, \textvisiblespace G}\\
        & $\mathbf{v}^{44,128}$ (22.7\%) & \texttt{\textvisiblespace A, \textvisiblespace C, \textvisiblespace E, \textvisiblespace B, \textvisiblespace S, \textvisiblespace G, \textvisiblespace F, \textvisiblespace P, \textvisiblespace D, \textvisiblespace K}\\
        & $\mathbf{v}^{38,4174}$ (20.2\%) & \texttt{\textvisiblespace C, \textvisiblespace c, C, \textvisiblespace B, \textvisiblespace D, \textvisiblespace P, \textvisiblespace F, \textvisiblespace T, \textvisiblespace L, \textvisiblespace S}\\
        & $\mathbf{v}^{37,423}$ (91.8\%) & \texttt{\textvisiblespace C, \textvisiblespace c, C, \textvisiblespace B, \textvisiblespace D, \textvisiblespace F, \textvisiblespace P, \textvisiblespace T, \textvisiblespace L, \textvisiblespace G}\\
        & $\mathbf{v}^{37,2966}$ (66.2\%) & \texttt{\textvisiblespace C, \textvisiblespace c, C, \textvisiblespace B, \textvisiblespace D, \textvisiblespace P, \textvisiblespace F, \textvisiblespace T, \textvisiblespace G, \textvisiblespace L}\\
    \bottomrule[1.5pt]
    \end{tabular}
    }
    \caption{The new top-10 tokens of each updated value vector for each GPT2 model.}
    \label{tab:new_value_unembed_more}
\end{table}
After directly updating each identified value vector from MLP, we further unembedded them and selected the top-10 tokens with the highest probability. Table~\ref{tab:new_value_unembed_more} shows that those top-10 tokens changes from original \texttt{`A'} to other correct choice tokens, i.e., \texttt{B, C, D, E}.

\section{Ablation Study of Different $\lambda_2$ in Eq.~\ref{eq:mlp_update}}\label{app:mlp_ablate}
We conduct an ablation study by changing $\lambda_2$ from 8 to 2 in Eq.~\ref{eq:mlp_update}, and evaluate the classification accuracy of each Infer. and Eva. dataset using GPT2 family. Table~\ref{tab:fix_mlp_success_rate_lambda_8},\ref{tab:fix_mlp_success_rate_lambda_7},\ref{tab:fix_mlp_success_rate_lambda_6},\ref{tab:fix_mlp_success_rate_lambda_5},\ref{tab:fix_mlp_success_rate_lambda_4},\ref{tab:fix_mlp_success_rate_lambda_3},\ref{tab:fix_mlp_success_rate_lambda_2} show the ablation study results under different $\lambda_2$. We can find that the classification accuracy shows a decreasing trend with $\lambda_2$ from 8 to 2, which indicates that the effect of direct value vector update using Eq.~\ref{eq:mlp_update} decreases.

\begin{table*}[ht]
    
    \centering
    \resizebox{0.99\textwidth}{!}{
    \begin{tabular}{l|l|cc|cc|cc|cc|cc}
    \toprule[1.5pt]
         \multirow{2}{*}{Model} & \multirow{2}{*}{Vector} & \multicolumn{2}{c|}{IOI (2)} & \multicolumn{2}{c|}{LD (3)} & \multicolumn{2}{c|}{Greater (4)} & \multicolumn{2}{c|}{ARC (4)} & \multicolumn{2}{c}{CSQA (5)}\\
         & & Infer. & Eva. & Infer. & Eva. & Infer. & Eva. & Infer. & Eva.& Infer. & Eva.\\\midrule
        GPT2-Small& $\mathbf{v}^{9,788}$ (50.0\%)& 100.0 & 100.0 & 100.0 & 100.0& 100.0 & 100.0 &100.0 & 100.0 &100.0 & 100.0\\\midrule
        GPT2-Medium& $\mathbf{v}^{20,1731}$ (38.2\%) & 33.6 & 33.7 & 81.8 & 50.0 & 0.0 & 0.0&51.2 &51.6 &22.4 &23.1\\\midrule
        \multirow{2}{*}{GPT2-Large}& $\mathbf{v}^{34,2103}$ (70.1\%) &71.2 &63.5 &55.6 &57.7 &32.0 & 96.2&49.7 &50.5 &52.9 &49.0\\
        & $\mathbf{v}^{34,4178}$ (55.4\%) & 25.1 &21.2 &12.2 & 19.2&33.2 & 93.3&45.2 & 46.2&55.1 &57.7\\\midrule
        \multirow{5}{*}{GPT2-XL}& $\mathbf{v}^{44,2995}$ (27.5\%) & 24.8 &36.3 &100.0 & 100.0&99.2 &100.0 &93.3 &92.3 &92.7 &94.2\\
        & $\mathbf{v}^{44,128}$ (22.7\%) & 67.8 &64.8 &96.8 &100.0 &100.0 &100.0 &91.1 &90.8 &92.5 &95.2\\
        & $\mathbf{v}^{38,4174}$ (20.2\%) &11.4 & 16.5&38.7 &23.1 &44.6 &75.0 &21.0 &26.2 &7.5 &7.7\\
        & $\mathbf{v}^{37,423}$ (91.8\%) &54.3 &60.4 &100.0 &100.0 &93.5 &100.0 &76.9 &81.5 &39.5 &36.5\\
        & $\mathbf{v}^{37,2966}$ (66.2\%) & 34.0&31.9 &100.0 &100.0 &77.2 &97.1 &73.7 &76.9 &55.2 &60.6\\
    \bottomrule[1.5pt]
    \end{tabular}
    }
    \caption{The classification accuracy of each MCQ inference (Infer.) and evaluation (Eva.) dataset after updating the identified value vectors for each GPT2 model using $\lambda_2=8$.}
    \label{tab:fix_mlp_success_rate_lambda_8}
\end{table*}

\begin{table*}[ht]
    
    \centering
    \resizebox{0.99\textwidth}{!}{
    \begin{tabular}{l|l|cc|cc|cc|cc|cc}
    \toprule[1.5pt]
         \multirow{2}{*}{Model} & \multirow{2}{*}{Vector} & \multicolumn{2}{c|}{IOI (2)} & \multicolumn{2}{c|}{LD (3)} & \multicolumn{2}{c|}{Greater (4)} & \multicolumn{2}{c|}{ARC (4)} & \multicolumn{2}{c}{CSQA (5)}\\
         & & Infer. & Eva. & Infer. & Eva. & Infer. & Eva. & Infer. & Eva.& Infer. & Eva.\\\midrule
        \multirow{3}{*}{GPT2-Small} & $\mathbf{v}^{9,1853}$ (100\%) & 99.5 & 100.0 & 100.0 & 100.0 & 100.0 & 100.0& 99.8& 100.0 &100.0 &100.0\\
        & $\mathbf{v}^{9,2859}$ (61.8\%) & 22.7 & 26.9 & 100.0 &100.0 & 37.8 & 69.2 &89.2 & 96.2 & 45.8& 41.0\\
        & $\mathbf{v}^{9,788}$ (50.0\%)& 100.0 & 100.0 & 100.0 & 100.0& 100.0 & 100.0 &100.0 & 100.0 &100.0 & 100.0\\\midrule
        \multirow{2}{*}{GPT2-Medium} & $\mathbf{v}^{20,3713}$ (79.3\%) & 97.7 & 95.2 &90.6 & 73.1 &17.3 & 47.1 &43.0 &52.7 &14.2 & 8.8\\
        & $\mathbf{v}^{20,1731}$ (38.2\%)& 27.6 & 26.9 & 61.2 & 30.8 & 0.0 & 0.0&41.4 &47.3 &13.8 &8.8\\\midrule
        \multirow{3}{*}{GPT2-Large} & $\mathbf{v}^{34,1541}$ (100\%) & 100.0 &100.0 &100.0 &100.0 &100.0 & 100.0&95.7 &95.6 & 99.2&99.0\\
        & $\mathbf{v}^{34,2103}$ (70.1\%) &57.6 &54.8 &40.6 &46.2 &26.6 & 85.6&38.6 &38.5 &42.6 &39.4\\
        & $\mathbf{v}^{34,4178}$ (55.4\%) & 16.8 &16.3 &2.2 & 3.8&27.0 & 85.6&35.5 & 37.4&45.5 &45.2\\\midrule
        \multirow{7}{*}{GPT2-XL} & $\mathbf{v}^{44,4967}$ (98.0\%) & 97.4 & 96.7&96.8 & 92.3&99.8 &100.0 &87.6 &90.8 &91.6 &92.3\\
        & $\mathbf{v}^{44,2995}$ (27.5\%) & 18.6 &30.8 &100.0 & 100.0&97.7 &100.0 &91.9 &92.3 &86.9 &89.4\\
        & $\mathbf{v}^{44,128}$ (22.7\%) & 59.6 &59.3 &93.5 &92.3 &100.0 &100.0 &90.0 &90.8 &89.7 &92.3\\
        & $\mathbf{v}^{38,4191}$ (100\%) & 99.4 &100.0 &100.0 & 100.0&88.0 &100.0 &96.1 &95.4 &92.6 &89.4\\
        & $\mathbf{v}^{38,4174}$ (20.2\%) &6.7 & 7.7&38.7 &23.1 &37.8 &60.6 &16.3 &18.5 &4.7 &4.8\\
        & $\mathbf{v}^{37,423}$ (91.8\%) &38.2 &41.8 &100.0 &100.0 &86.9 &100.0 &67.3 &70.8 &30.1 &29.8\\
        & $\mathbf{v}^{37,2966}$ (66.2\%) & 22.3&26.4 &93.5 &84.6 &71.3 &96.2 &65.7 &64.6 &44.6 &48.1\\
    \bottomrule[1.5pt]
    \end{tabular}
    }
    \caption{The classification accuracy of each MCQ inference (Infer.) and evaluation (Eva.) dataset after updating the identified value vectors for each GPT2 model using $\lambda_2=7$.}
    \label{tab:fix_mlp_success_rate_lambda_7}
    % \vspace{-4.5mm}
\end{table*}

\begin{table*}[ht]
    
    \centering
    \resizebox{0.99\textwidth}{!}{
    \begin{tabular}{l|l|cc|cc|cc|cc|cc}
    \toprule[1.5pt]
         \multirow{2}{*}{Model} & \multirow{2}{*}{Vector} & \multicolumn{2}{c|}{IOI (2)} & \multicolumn{2}{c|}{LD (3)} & \multicolumn{2}{c|}{Greater (4)} & \multicolumn{2}{c|}{ARC (4)} & \multicolumn{2}{c}{CSQA (5)}\\
         & & Infer. & Eva. & Infer. & Eva. & Infer. & Eva. & Infer. & Eva.& Infer. & Eva.\\\midrule
        \multirow{3}{*}{GPT2-Small} & $\mathbf{v}^{9,1853}$ (100\%) & 96.1 & 96.2 & 100.0 & 100.0 & 94.8 & 100.0& 98.7& 98.1 &99.4 &97.4\\
        & $\mathbf{v}^{9,2859}$ (61.8\%) & 7.6 & 7.7 & 92.1 &100.0 & 20.1 & 38.5 &79.8 & 84.6 & 25.6& 25.6\\
        & $\mathbf{v}^{9,788}$ (50.0\%)& 100.0 & 100.0 & 100.0 & 100.0& 100.0 & 100.0 &100.0 & 100.0 &100.0 & 100.0\\\midrule
        \multirow{2}{*}{GPT2-Medium} & $\mathbf{v}^{20,3713}$ (79.3\%) &92.5 & 90.4 &77.6 & 26.9 &7.8 & 26.0 &29.3 &37.4 &4.9 & 2.2\\
        & $\mathbf{v}^{20,1731} $ (38.2\%)& 22.0 & 20.2 & 31.2 & 11.5 & 0.0 & 0.0&29.7 &36.3 &7.1 &5.5\\\midrule
        \multirow{3}{*}{GPT2-Large} & $\mathbf{v}^{34,1541}$ (100\%) & 100.0 &100.0 &100.0 &100.0 &100.0 & 100.0&93.1 &91.2 & 97.8&97.1\\
        & $\mathbf{v}^{34,2103} $ (70.1\%)&39.8 &37.5 &20.6 &46.2 &14.8 & 51.9&28.7 &29.7 &30.8 &26.9\\
        & $\mathbf{v}^{34,4178}$ (55.4\%) & 9.8 &7.7 &0.6 & 0.0&15.7 & 51.0&26.1 & 28.6&35.0 &31.7\\\midrule
        \multirow{7}{*}{GPT2-XL} & $\mathbf{v}^{44,4967}$ (98.0\%) & 95.2 & 95.6&93.5 & 84.6&99.4 &100.0 &84.2 &90.8 &85.5 &87.5\\
        & $\mathbf{v}^{44,2995}$ (27.5\%) & 12.9 &19.8 &96.8 & 92.3&93.1 &100.0 &87.9 &92.3 &76.6 &85.6\\
        & $\mathbf{v}^{44,128}$ (22.7\%) & 50.7 &50.5 &90.3 &84.6 &99.8 &100.0 &87.6 &89.2 &84.8 &87.5\\
        & $\mathbf{v}^{38,4191}$ (100\%) & 96.9 &98.9 &100.0 & 100.0&78.1 &100.0 &95.1 &95.4 &82.5 &83.7\\
        & $\mathbf{v}^{38,4174}$ (20.2\%) &3.2 & 5.5&35.5 &23.1 &29.9 &47.1 &13.0 &16.9 &2.7 &1.0\\
        & $\mathbf{v}^{37,423}$ (91.8\%) &23.2 &27.5 &100.0 &100.0 &80.3 &100.0 &57.8 &60.0 &20.5 &17.3\\
        & $\mathbf{v}^{37,2966}$ (66.2\%) & 12.5&15.4 &77.4 &76.9 &60.7 &92.3 &56.0 &56.9 &32.4 &29.8\\
    \bottomrule[1.5pt]
    \end{tabular}
    }
    \caption{The classification accuracy of each MCQ inference (Infer.) and evaluation (Eva.) dataset after updating the identified value vectors for each GPT2 model using $\lambda_2=6$.}
    \label{tab:fix_mlp_success_rate_lambda_6}
\end{table*}

\begin{table*}[ht]
    
    \centering
    \resizebox{0.99\textwidth}{!}{
    \begin{tabular}{l|l|cc|cc|cc|cc|cc}
    \toprule[1.5pt]
         \multirow{2}{*}{Model} & \multirow{2}{*}{Vector} & \multicolumn{2}{c|}{IOI (2)} & \multicolumn{2}{c|}{LD (3)} & \multicolumn{2}{c|}{Greater (4)} & \multicolumn{2}{c|}{ARC (4)} & \multicolumn{2}{c}{CSQA (5)}\\
         & & Infer. & Eva. & Infer. & Eva. & Infer. & Eva. & Infer. & Eva.& Infer. & Eva.\\\midrule
        \multirow{3}{*}{GPT2-Small} & $\mathbf{v}^{9,1853}$ (100\%) & 76.8 & 82.7 & 100.0 & 100.0 & 69.2 & 100.0& 93.0& 96.2 &96.1 &94.9\\
        & $\mathbf{v}^{9,2859}$ (61.8\%) & 1.2 & 3.8 & 48.2 &69.2 & 4.5 & 12.8 &63.2 & 69.2 & 12.3& 12.8\\
        & $\mathbf{v}^{9,788}$ (50.0\%)& 100.0 & 100.0 & 100.0 & 100.0& 100.0 & 100.0 &100.0 & 100.0 &100.0 & 100.0\\\midrule
        \multirow{2}{*}{GPT2-Medium} & $\mathbf{v}^{20,3713}$ (79.3\%) &79.8 & 76.9 &50.0 & 0.0 &0.9 &1.9 &18.9 &24.2 &0.8 & 0.0\\
        & $\mathbf{v}^{20,1731} $ (38.2\%)& 15.7 & 15.4 & 10.0 & 0.0 & 0.0 & 0.0&20.2 &18.7 &3.1 &2.2\\\midrule
        \multirow{3}{*}{GPT2-Large} & $\mathbf{v}^{34,1541}$ (100\%) & 100.0 &100.0 &100.0 &100.0 &100.0 & 100.0&87.6 &83.5 & 95.3&95.2\\
        & $\mathbf{v}^{34,2103} $ (70.1\%)&23.4 &17.3 &7.8 &26.9 &3.6 & 10.6&20.2 &13.2 &21.2 &14.4\\
        & $\mathbf{v}^{34,4178}$ (55.4\%) & 4.6 &2.9 &0.0 & 0.0&5.4 & 20.2&17.4 & 16.5&25.1 &17.3\\\midrule
        \multirow{7}{*}{GPT2-XL} & $\mathbf{v}^{44,4967}$ (98.0\%) & 88.6 & 90.1&77.4 & 69.2&95.8 &100.0 &77.6 &84.6 &76.0 &78.8\\
        & $\mathbf{v}^{44,2995}$ (27.5\%) & 7.2 &15.4 &90.3 & 84.6&86.2 &98.1 &82.7 &84.6 &61.0 &69.2\\
        & $\mathbf{v}^{44,128}$ (22.7\%) & 38.9 &41.8 &83.9 &84.6 &98.5 &100.0 &82.5 &86.2 &74.8 &76.0\\
        & $\mathbf{v}^{38,4191}$ (100\%) & 85.0 &87.9 &100.0 & 100.0&68.4 &99.0 &88.1 &87.7 &64.4 &67.3\\
        & $\mathbf{v}^{38,4174}$ (20.2\%) &1.3 & 4.4&32.3 &15.4 &21.1 &35.6 &9.3 &13.8 &1.7 &0.0\\
        & $\mathbf{v}^{37,423}$ (91.8\%) &12.5 &17.6 &96.8 &100.0 &69.8 &95.2 &47.8 &47.7 &13.5 &13.5\\
        & $\mathbf{v}^{37,2966}$ (66.2\%) & 5.6&8.8 &64.5 &61.5 &49.2 &73.1 &43.4 &40.0 &21.2 &20.2\\
    \bottomrule[1.5pt]
    \end{tabular}
    }
    \caption{The classification accuracy of each MCQ inference (Infer.) and evaluation (Eva.) dataset after updating the identified value vectors for each GPT2 model using $\lambda_2=5$.}
    \label{tab:fix_mlp_success_rate_lambda_5}
\end{table*}

\begin{table*}[ht]
    
    \centering
    \resizebox{0.99\textwidth}{!}{
    \begin{tabular}{l|l|cc|cc|cc|cc|cc}
    \toprule[1.5pt]
         \multirow{2}{*}{Model} & \multirow{2}{*}{Vector} & \multicolumn{2}{c|}{IOI (2)} & \multicolumn{2}{c|}{LD (3)} & \multicolumn{2}{c|}{Greater (4)} & \multicolumn{2}{c|}{ARC (4)} & \multicolumn{2}{c}{CSQA (5)}\\
         & & Infer. & Eva. & Infer. & Eva. & Infer. & Eva. & Infer. & Eva.& Infer. & Eva.\\\midrule
        \multirow{3}{*}{GPT2-Small} & $\mathbf{v}^{9,1853}$ (100\%) & 30.7 & 36.5 & 58.8 & 69.2 & 26.3 & 59.0& 63.9& 73.1 &64.9 &69.2\\
        & $\mathbf{v}^{9,2859}$ (61.8\%) & 0.0 & 0.0 & 3.5 &0.0 & 0.0 & 0.0 &35.0 & 44.2 & 2.3& 2.6\\
        & $\mathbf{v}^{9,788}$ (50.0\%)& 99.5 & 100.0 & 100.0 & 100.0& 100.0 & 100.0 &97.8 & 100.0 &99.7 & 100.0\\\midrule
        \multirow{2}{*}{GPT2-Medium} & $\mathbf{v}^{20,3713}$ (79.3\%) &59.4 & 55.8 &16.5 & 0.0 &0.0 &0.0 &7.6 &9.9 &0.1 & 0.0\\
        & $\mathbf{v}^{20,1731} $ (38.2\%)& 11.1 & 9.6 & 0.6 & 0.0 & 0.0 & 0.0&10.4 &13.2 &1.0 &0.0\\\midrule
        \multirow{3}{*}{GPT2-Large} & $\mathbf{v}^{34,1541}$ (100\%) & 100.0 &100.0 &100.0 &100.0 &98.2 & 100.0&80.7 &79.1 & 86.8&89.4\\
        & $\mathbf{v}^{34,2103} $ (70.1\%)&10.3 &4.8 &1.1 &0.0 &0.1 & 0.0&10.9 &8.8 &12.6 &5.8\\
        & $\mathbf{v}^{34,4178}$ (55.4\%) & 2.2 &1.9 &0.0 & 0.0&0.6 & 1.0&12.3 & 9.9&15.9 &10.6\\\midrule
        \multirow{7}{*}{GPT2-XL} & $\mathbf{v}^{44,4967}$ (98.0\%) & 74.6 & 76.9&61.3 & 38.5&85.7 &100.0 &68.0 &76.9 &57.4 &62.5\\
        & $\mathbf{v}^{44,2995}$ (27.5\%) & 4.1 &9.9 &77.4 & 61.5&73.4 &96.2 &73.9 &75.4 &42.9 &50.0\\
        & $\mathbf{v}^{44,128}$ (22.7\%) & 26.3 &31.9 &71.0 &61.5 &92.1 &99.0 &75.7 &78.5 &58.9 &65.4\\
        & $\mathbf{v}^{38,4191}$ (100\%) & 58.9 &57.1 &71.0 & 61.5&58.1 &90.4 &75.5 &75.4 &43.2 &45.2\\
        & $\mathbf{v}^{38,4174}$ (20.2\%) &0.5 & 2.2&32.3 &15.4 &13.1 &22.1 &6.8 &10.8 &1.2 &0.0\\
        & $\mathbf{v}^{37,423}$ (91.8\%) &4.4 &3.3 &80.6 &76.9 &54.9 &76.0 &34.7 &38.5 &7.5 &4.8\\
        & $\mathbf{v}^{37,2966}$ (66.2\%) & 1.9&4.4 &51.6 &38.5 &33.5 &54.8 &32.2 &30.8 &11.1 &7.7\\
    \bottomrule[1.5pt]
    \end{tabular}
    }
    \caption{The classification accuracy of each MCQ inference (Infer.) and evaluation (Eva.) dataset after updating the identified value vectors for each GPT2 model using $\lambda_2=4$.}
    \label{tab:fix_mlp_success_rate_lambda_4}
\end{table*}

\begin{table*}[ht]
    
    \centering
    \resizebox{0.99\textwidth}{!}{
    \begin{tabular}{l|l|cc|cc|cc|cc|cc}
    \toprule[1.5pt]
         \multirow{2}{*}{Model} & \multirow{2}{*}{Vector} & \multicolumn{2}{c|}{IOI (2)} & \multicolumn{2}{c|}{LD (3)} & \multicolumn{2}{c|}{Greater (4)} & \multicolumn{2}{c|}{ARC (4)} & \multicolumn{2}{c}{CSQA (5)}\\
         & & Infer. & Eva. & Infer. & Eva. & Infer. & Eva. & Infer. & Eva.& Infer. & Eva.\\\midrule
        \multirow{3}{*}{GPT2-Small} & $\mathbf{v}^{9,1853}$ (100\%) & 1.7 & 1.9 & 0.0 & 0.0 & 2.1 & 10.3& 16.8& 23.1 &14.3 &15.4\\
        & $\mathbf{v}^{9,2859}$ (61.8\%) & 0.0 & 0.0 & 0.0 &0.0 & 0.0 & 0.0 &8.5 & 9.6 & 0.0& 0.0\\
        & $\mathbf{v}^{9,788}$ (50.0\%)& 67.6 & 71.2 & 71.1 & 76.9& 98.3 & 100.0 &70.6 & 73.1 &92.9 & 97.4\\\midrule
        \multirow{2}{*}{GPT2-Medium} & $\mathbf{v}^{20,3713}$ (79.3\%) &34.7 & 31.7 &0.0 & 0.0 &0.0 &0.0 &1.5 &3.3 &0.0 & 0.0\\
        & $\mathbf{v}^{20,1731} $ (38.2\%)& 8.1 & 4.8 & 0.0 & 0.0 & 0.0 & 0.0&4.8 &6.6 &0.4 &0.0\\\midrule
        \multirow{3}{*}{GPT2-Large} & $\mathbf{v}^{34,1541}$ (100\%) & 100.0 &100.0 &95.0 &100.0 &67.7 & 100.0&65.1 &68.1 & 70.4&72.1\\
        & $\mathbf{v}^{34,2103} (70.1\%)$ &3.1 &1.9 &0.0 &0.0 &0.0 & 0.0&6.1 &3.3 &5.8 &4.8\\
        & $\mathbf{v}^{34,4178}$ (55.4\%) & 0.9 &0.0 &0.0 & 0.0&0.0 & 0.0&6.1 & 4.4&6.5 &5.8\\\midrule
        \multirow{7}{*}{GPT2-XL} & $\mathbf{v}^{44,4967}$ (98.0\%) & 51.2 & 58.2&61.3 & 38.5&64.2 &91.3 &52.0 &53.8 &33.3 &34.6\\
        & $\mathbf{v}^{44,2995}$ (27.5\%) & 1.8 &4.4 &61.3 & 46.2&49.8 &78.8 &56.4 &58.5 &24.0 &24.0\\
        & $\mathbf{v}^{44,128}$ (22.7\%) & 12.7 &17.6 &54.8 &53.8 &76.6 &94.2 &61.8 &64.6 &40.4 &42.3\\
        & $\mathbf{v}^{38,4191}$ (100\%) & 21.7 &27.5 &58.1 & 38.5&38.1 &63.5 &54.1 &63.1 &19.7 &21.2\\
        & $\mathbf{v}^{38,4174}$ (20.2\%) &0.1 & 0.0&29.0 &7.7 &6.2 &13.5 &4.4 &6.2 &0.8 &0.0\\
        & $\mathbf{v}^{37,423}$ (91.8\%) &1.3 &2.2 &67.7 &53.8 &34.7 &52.9 &22.2 &24.6 &3.7 &3.8\\
        & $\mathbf{v}^{37,2966}$ (66.2\%) & 0.5&1.1 &48.4 &30.8 &19.6 &37.5 &22.2 &18.5 &6.6 &5.8\\
    \bottomrule[1.5pt]
    \end{tabular}
    }
    \caption{The classification accuracy of each MCQ inference (Infer.) and evaluation (Eva.) dataset after updating the identified value vectors for each GPT2 model using $\lambda_2=3$.}
    \label{tab:fix_mlp_success_rate_lambda_3}
\end{table*}

\begin{table*}[ht]
    
    \centering
    \resizebox{0.99\textwidth}{!}{
    \begin{tabular}{l|l|cc|cc|cc|cc|cc}
    \toprule[1.5pt]
         \multirow{2}{*}{Model} & \multirow{2}{*}{Vector} & \multicolumn{2}{c|}{IOI (2)} & \multicolumn{2}{c|}{LD (3)} & \multicolumn{2}{c|}{Greater (4)} & \multicolumn{2}{c|}{ARC (4)} & \multicolumn{2}{c}{CSQA (5)}\\
         & & Infer. & Eva. & Infer. & Eva. & Infer. & Eva. & Infer. & Eva.& Infer. & Eva.\\\midrule
        \multirow{3}{*}{GPT2-Small} & $\mathbf{v}^{9,1853}$ (100\%) & 0.0 & 0.0 & 0.0 & 0.0 & 0.0 & 0.0& 1.3& 1.9 &1.0 &0.0\\
        & $\mathbf{v}^{9,2859}$ (61.8\%) & 0.0 & 0.0 & 0.0 &0.0 & 0.0 & 0.0 &0.9 & 1.9 & 0.0& 0.0\\
        & $\mathbf{v}^{9,788}$ (50.0\%)& 5.4 & 7.7 & 0.0 & 0.0& 24.2 & 53.8 &15.5 & 19.2 &26.9 & 30.8\\\midrule
        \multirow{2}{*}{GPT2-Medium} & $\mathbf{v}^{20,3713}$ (79.3\%) &15.6 & 16.3 &0.0 & 0.0 &0.0 &0.0 &0.3 &0.0 &0.0 & 0.0\\
        & $\mathbf{v}^{20,1731} $ (38.2\%)& 4.2 & 1.9 & 0.0 & 0.0 & 0.0 & 0.0&1.7 &4.3 &0.0 &0.0\\\midrule
        \multirow{3}{*}{GPT2-Large} & $\mathbf{v}^{34,1541}$ (100\%) & 99.9 &100.0 &66.1 &69.2 &34.4 & 100.0&37.5 &37.4 & 39.5&34.6\\
        & $\mathbf{v}^{34,2103} $ (70.1\%)&0.6 &1.0 &0.0 &0.0 &0.0 & 0.0&2.3 &2.2 &1.8 &1.9\\
        & $\mathbf{v}^{34,4178}$ (55.4\%) & 0.1 &0.0 &0.0 & 0.0&0.0 & 0.0&2.1 & 1.1&2.7 &2.9\\\midrule
        \multirow{7}{*}{GPT2-XL} & $\mathbf{v}^{44,4967}$ (98.0\%) & 18.5 & 17.6&48.4 & 23.1&33.3 &62.5 &31.7 &33.8 &13.8 &14.4\\
        & $\mathbf{v}^{44,2995}$ (27.5\%) & 0.6 &2.2 &48.4 & 30.8&25.4 &49.0 &36.1 &35.4 &9.0 &5.8\\
        & $\mathbf{v}^{44,128}$ (22.7\%) & 4.0 &7.7 &48.4 &38.5 &48.9 &76.9 &43.1 &38.5 &18.9 &20.2\\
        & $\mathbf{v}^{38,4191}$ (100\%) & 4.3 &6.6 &41.9 & 23.1&16.1 &28.8 &28.9&29.2 &6.5 &5.8\\
        & $\mathbf{v}^{38,4174}$ (20.2\%) &0.0 & 0.0&22.6 &7.7 &2.8 &4.8 &2.5 &4.6 &0.5 &0.0\\
        & $\mathbf{v}^{37,423}$ (91.8\%) &0.4 &1.1 &48.4 &30.8 &15.6 &32.7 &12.3 &12.3 &1.5 &1.9\\
        & $\mathbf{v}^{37,2966}$ (66.2\%) & 0.1&0.0 &32.3 &15.4 &7.5 &15.4 &11.2 &15.4 &2.0 &0.0\\
    \bottomrule[1.5pt]
    \end{tabular}
    }
    \caption{The classification accuracy of each MCQ inference (Infer.) and evaluation (Eva.) dataset after updating the identified value vectors for each GPT2 model using $\lambda_2=2$.}
    \label{tab:fix_mlp_success_rate_lambda_2}
\end{table*}

\section{Top-1 Classification Accuracy Changes Using Attention Pattern Recalibration}\label{app:attn_recali}
We also evaluate the top-1 classification accuracy when the attention pattern recalibration is used to the identified attention head. In Table~\ref{tab:mitigate_attn_acc_more}, we can find that most top-1 classification accuracy is close to 0\%, which indicates that those identified attention heads play less important roles compared to specific MLP modules in each GPT2 model under the MCQ task setting.
\begin{table*}[ht]
    
    \centering
    \resizebox{0.99\textwidth}{!}{
    \begin{tabular}{l|l|cccccccccc}
    \toprule[1.5pt]
         \multirow{2}{*}{Model} & \multirow{2}{*}{Head} & \multicolumn{2}{c}{IOI (2)} & \multicolumn{2}{c}{LD (3)} & \multicolumn{2}{c}{Greater (4)} & \multicolumn{2}{c}{ARC (4)} & \multicolumn{2}{c}{CSQA (5)}\\
         & & Infer. & Eva. & Infer. & Eva. & Infer. & Eva. & Infer. & Eva.& Infer. & Eva.\\\midrule
        \multirow{2}{*}{GPT2-Small} & L8H1 & 0.0 & 0.0 & 0.0 & 0.0 & 0.0 & 0.0 & 0.0 & 2.04 & 0.0 & 0.0\\
        & L10H8 & 0.0 & 0.0 & 0.0 & 0.0 & 0.0 & 0.0 & 0.0 & 2.04 & 0.0 & 0.0\\\midrule
        \multirow{2}{*}{GPT2-Medium} & L18H12 & 92.47 & 90.72 & 21.76 & 27.78 & 3.39 & 3.16 & 24.33 & 21.69 & 1.67 & 1.25\\
        & L20H5 & 0.34 & 0.0 & 0.0 & 0.0 & 0.0 & 0.0 & 0.0 & 0.0 & 0.0 & 0.0\\\midrule
        \multirow{2}{*}{GPT2-Large} & L23H8 & 0.0 & 0.0 & 0.0 & 0.0 & 0.0 & 0.0 & 0.38 & 1.14 & 0.0 & 2.06\\
        & L30H0 & 0.0 & 0.0 & 0.0 & 0.0 & 0.0 & 0.0 & 0.0 & 0.0 & 0.0 & 0.0\\\midrule
        \multirow{2}{*}{GPT2-XL} & L31H9 & 0.0 & 0.0 & 9.68 & 0.0 & 0.57 & 0.0 & 0.53 & 0.0 & 0.12 & 0.0\\
        & L34H14 & 0.0 & 0.0 & 12.90 & 0.0 & 0.57 & 0.0 &  3.33& 1.67 & 0.46 & 0.0\\
    \bottomrule[1.5pt]
    \end{tabular}
    }
    \caption{The Top-1 classification accuracy changes after attention pattern recalibration for each GPT2 model across all datasets.}
    \label{tab:mitigate_attn_acc_more}
\end{table*}

\section{One-shot and Two-shot MCQ Classification Results}\label{app:few_shot}
\begin{figure*}[ht]
    \centering
    \includegraphics[width=\textwidth]{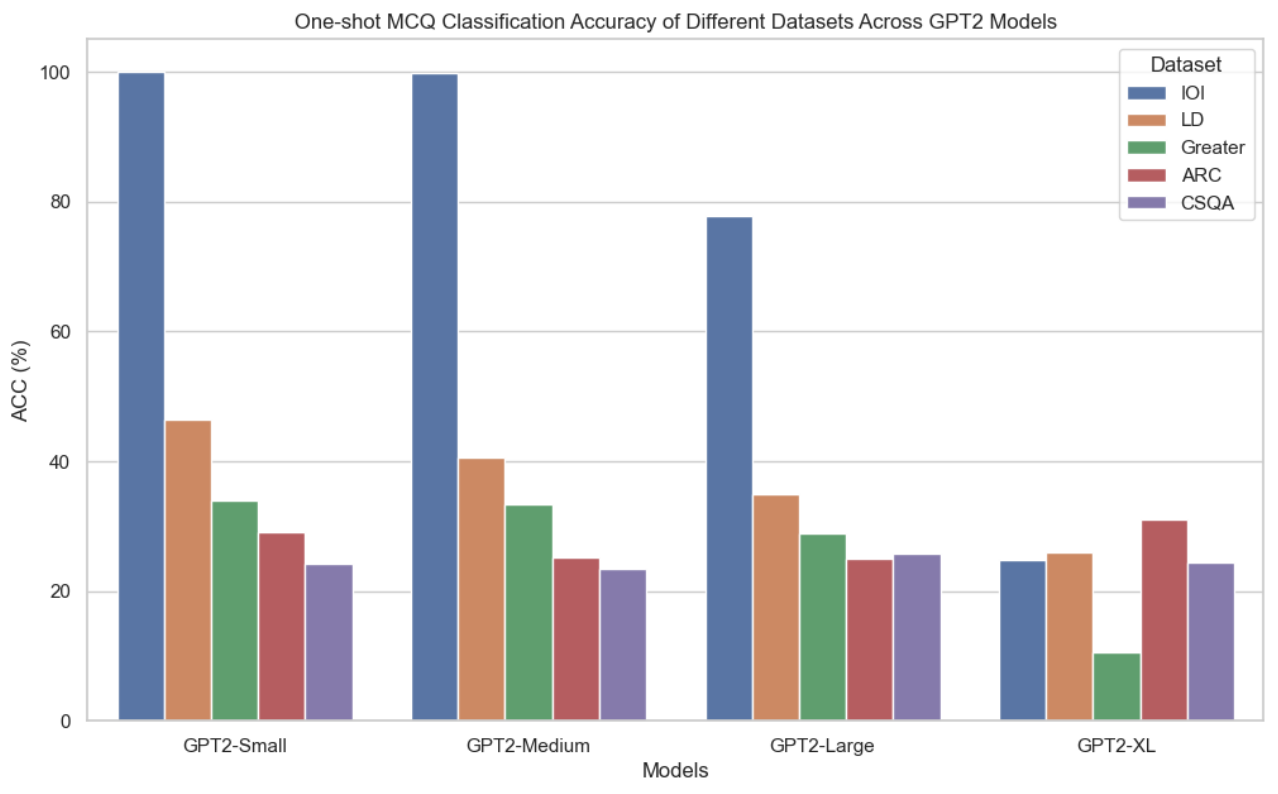}
    \caption{One-shot MCQ classification accuracy of different datasets across GPT2 family.}
    \label{fig:one_shot_acc}
\end{figure*}

\begin{figure*}[ht]
    \centering
    \includegraphics[width=\textwidth]{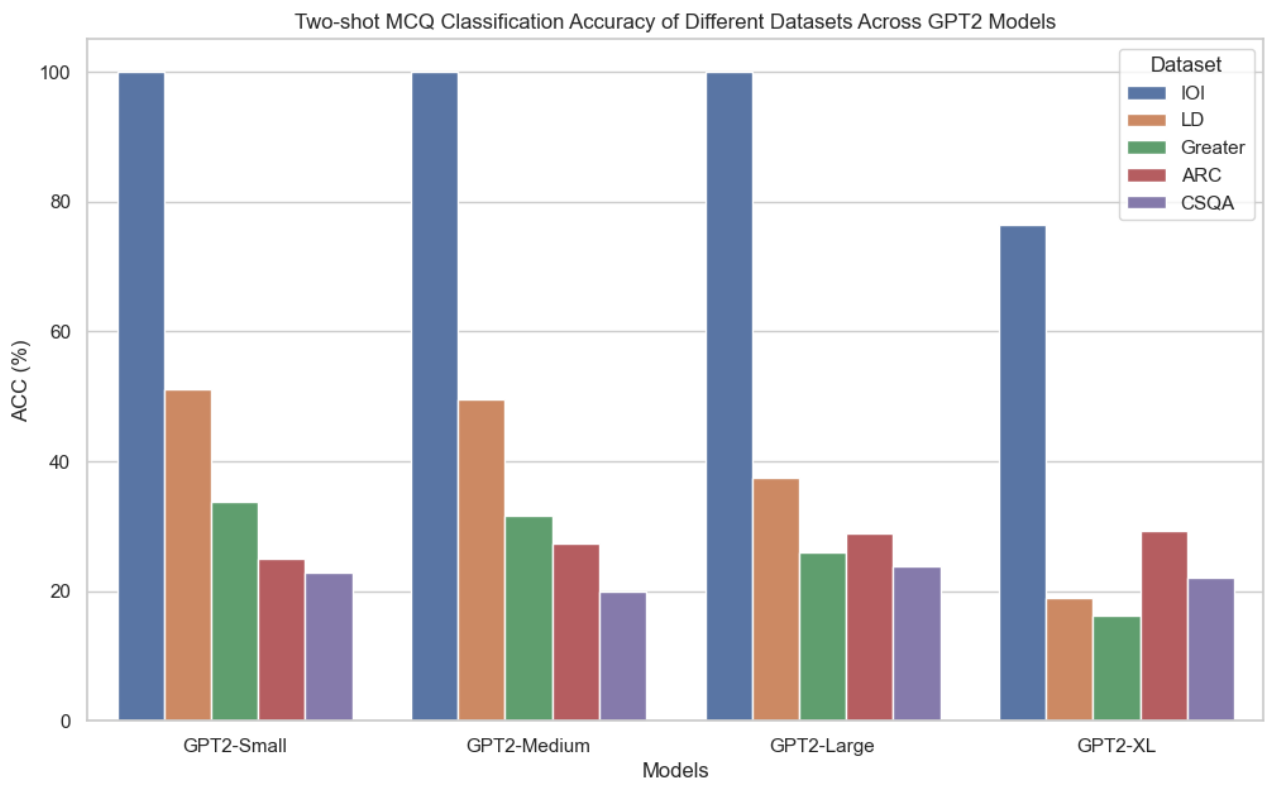}
    \caption{Two-shot MCQ classification accuracy of different datasets across GPT2 family.}
    \label{fig:two_shot_acc}
\end{figure*}

Based on the discussion in \S~\ref{s:discussion}, we further conduct an experiment to evaluate whether the anchored bias can be mitigated under the few-shot learning setting. Fig.~\ref{fig:one_shot_acc} and Fig.~\ref{fig:two_shot_acc} show that 2-shot leaning performs better than 1-shot setting across all datasets, especially for IOI dataset. However, GPT2-XL always struggle to predict higher accuracy across all datasets. This finding indicates that simple few-shot learning cannot mitigate anchored bias and a more comprehensive analysis is needed to investigate this bias.

\section{Damage to Updated GPT2 Family's Performance on IOI and Greater Datasets}\label{app:damage}
\begin{figure*}[ht]
    \centering
    \includegraphics[width=\textwidth]{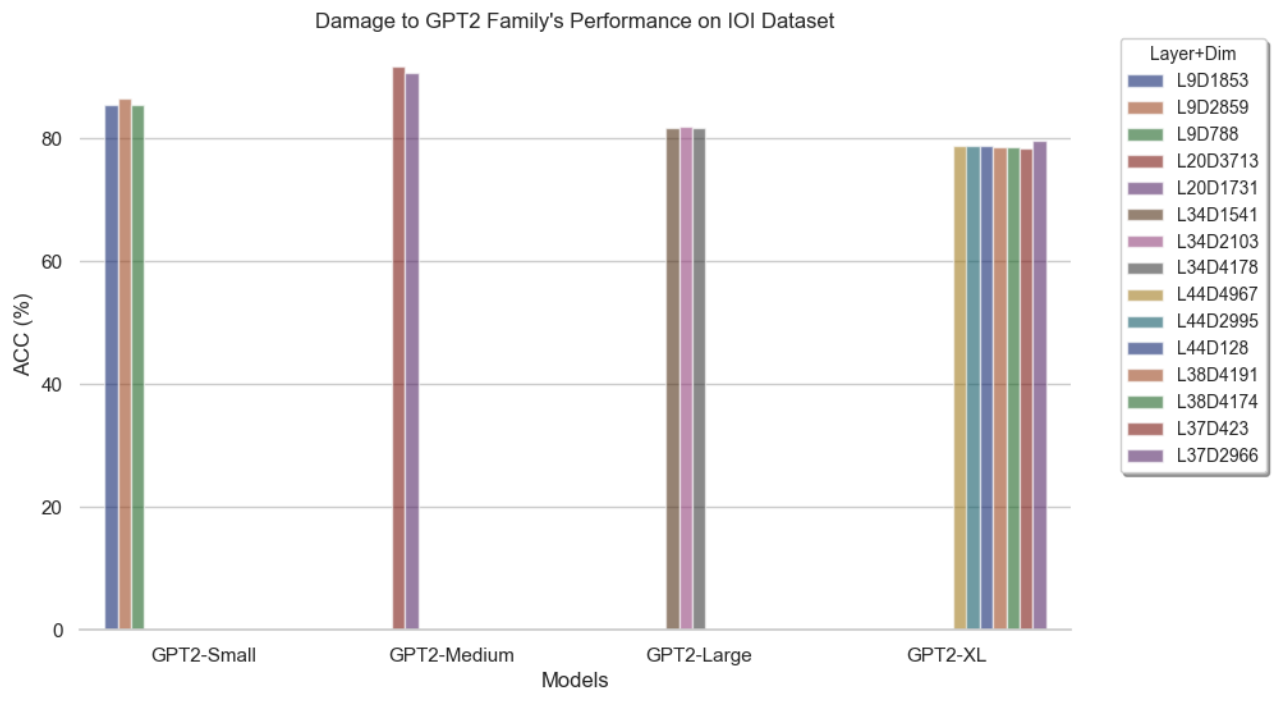}
    \caption{Damage to updated GPT2 family's classification accuracy on original IOI dataset using Eq.~\ref{eq:mlp_update} with $\lambda_2=8$.}
    \label{fig:damage_IOI}
\end{figure*}

\begin{figure*}[ht]
    \centering
    \includegraphics[width=\textwidth]{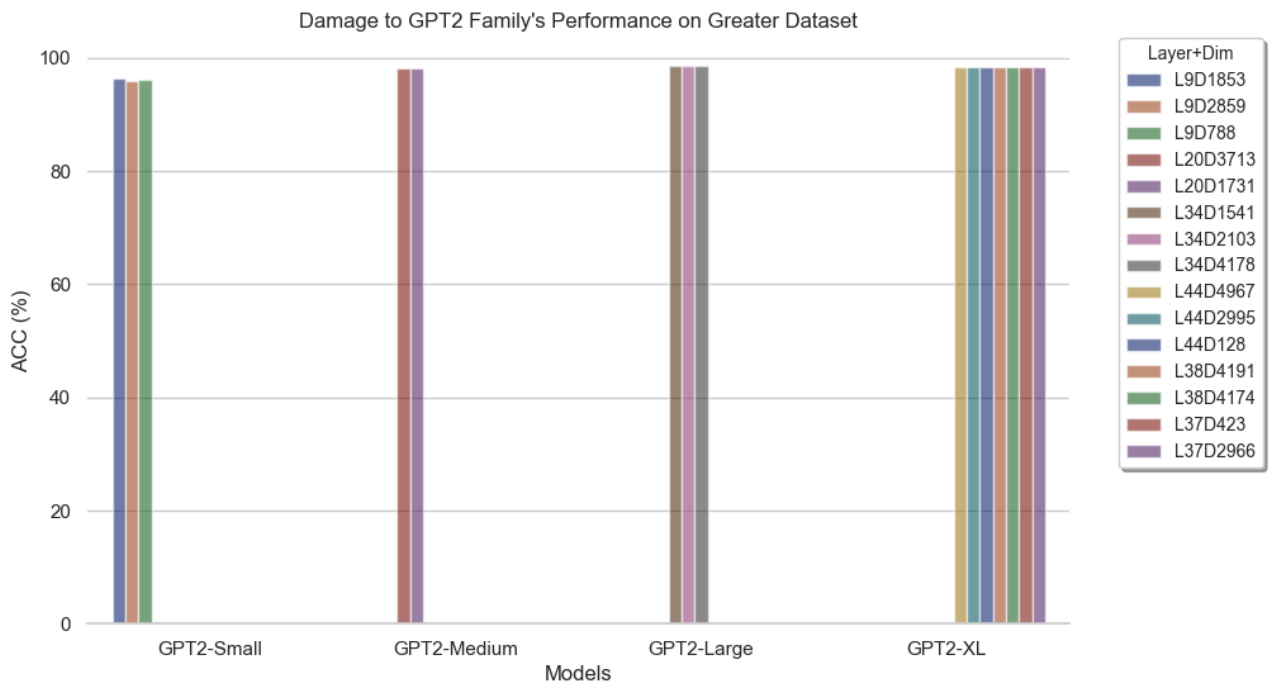}
    \caption{Damage to updated GPT2 family's classification accuracy on original Greater dataset using Eq.~\ref{eq:mlp_update} with $\lambda_2=8$.}
    \label{fig:damage_Greater}
\end{figure*}

Based on the discussion in \S~\ref{s:discussion}, we further evaluate the damage of direct value vector update from MLP for the general ability of GPT2 family. We choose original IOI and Greater-than datasets as they have been verified that GPT2 family works well~\citep{wang2022interpretability,merullo2023circuit,hanna2024does}. Fig.~\ref{fig:damage_IOI} and Fig.~\ref{fig:damage_Greater} show that classification accuracy is acceptable when a value vector with a specific layer and dimensionality is updated, which matches the findings from~\citet{gu-etal-2024-model}.

\section{MCQ Prompt Template with Random Words}\label{app:random_words}
In order to evaluate whether anchored bias is sensitive to the specific content of the input MCQ prompts, we built two random MCQ datasets. One of them contains randomly selected characters and we concatenate them as a word with length from 5 to 10, and each MCQ prompt is built based on ``\textit{Question: <Question sample> Answer Choices: <Multiple Choices> Answer:}''. For the random word dataset, we randomly select words from a word vocabulary\footnote{\url{https://www.mit.edu/~ecprice/wordlist.10000}}. Finally, each dataset has 80 samples with the number of choices from 2 to 5 (see Table~\ref{tab:random_MCQ_prompt_example}). As shown in Table~\ref{tab:random_MCQ_prompt}, anchored bias still happens across GPT2 family, especially for GPT2-Large and GPT2-XL. This finding confirms our findings that GPT2 family exhibit the anchored bias with significant regularity across various MCQ datasets.
\begin{table}[ht]
    
    \centering
    \resizebox{\columnwidth}{!}{
    \begin{tabular}{l|cccc}
    \toprule[1.5pt]
      Type & GPT2-Small & GPT2-Medium & GPT2-Large & GPT2-XL\\\midrule
      Random Characters & 70.0 & 69.0& 99.0& 96.0 \\
      Random Words & 65.0 & 82.0 & 95.0& 100.0\\
    \bottomrule[1.5pt]
    \end{tabular}
    }
    \caption{The percentage of anchored bias occurring when MCQ prompt template is replaced with random characters and random words.}
    \label{tab:random_MCQ_prompt}
\end{table}

\begin{table*}[ht]
    
    \centering
    \resizebox{0.89\textwidth}{!}{
    \begin{tabular}{p{1.5cm}|p{10cm}}
    \toprule[1.5pt]
      Random Characters & Question: cgryp rkodmajc ajwsqby jnqqaqcmsa agbyyasj ibngbxtn dazvqigre urbnmumw ltpjslayp ighfudgy hbaldde? Answer Choices: A: samdaheepw ltvlpeh B: rqqtxdgiyb rznosxhk djpsitdar ubjgq ioamje C: bkdjziiy Answer:\\\hline
      Random Words & Question: citysearch logical bidder discount kentucky forming rapid digit flash putting reid liechtenstein mate? Answer Choices: A: seven owner voluntary  B: clicking it  C: harassment beam firewire  D: run helpful  Answer:\\
    \bottomrule[1.5pt]
    \end{tabular}
    }
    \caption{The MCQ prompt template with random characters and random words.}
    \label{tab:random_MCQ_prompt_example}
\end{table*}

\section{Full Circuit of Anchored Bias of Each GPT2 Model}\label{app:gpt2_curcuit}
\begin{figure*}[ht]
    \centering
    \includegraphics[width=\textwidth]{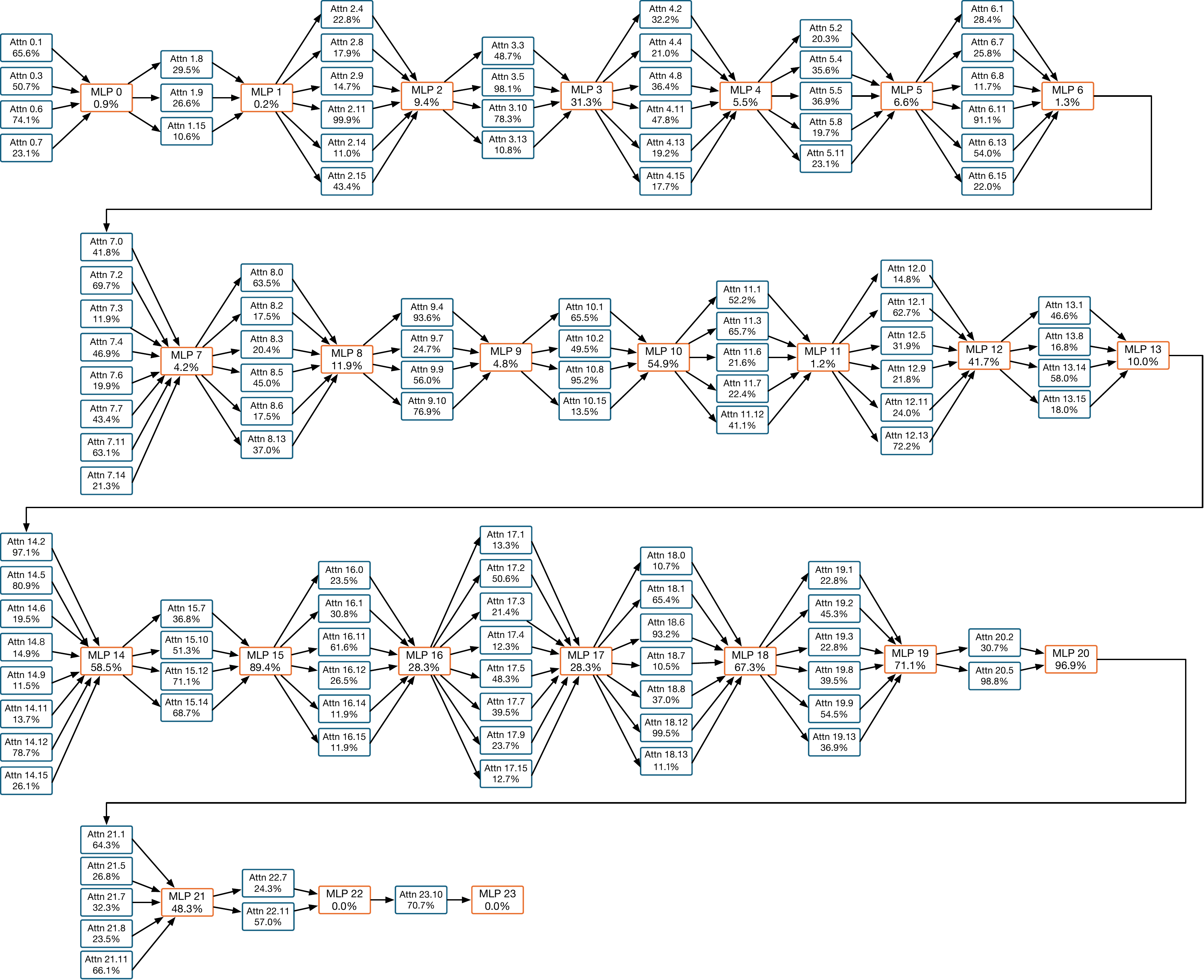}
    \caption{The full circuit of anchored bias for GPT2-Medium model, where each attention head and MLP module are selected when MLP and attention pattern logit difference threshold is larger than 4. The percentage within each module indicates the probability of anchored bias across different datasets for GPT2-Medium model when the threshold is larger than 4.}
    \label{fig:gpt2_medium_circuit}
\end{figure*}

\begin{figure*}[ht]
    \centering
    \includegraphics[width=\textwidth]{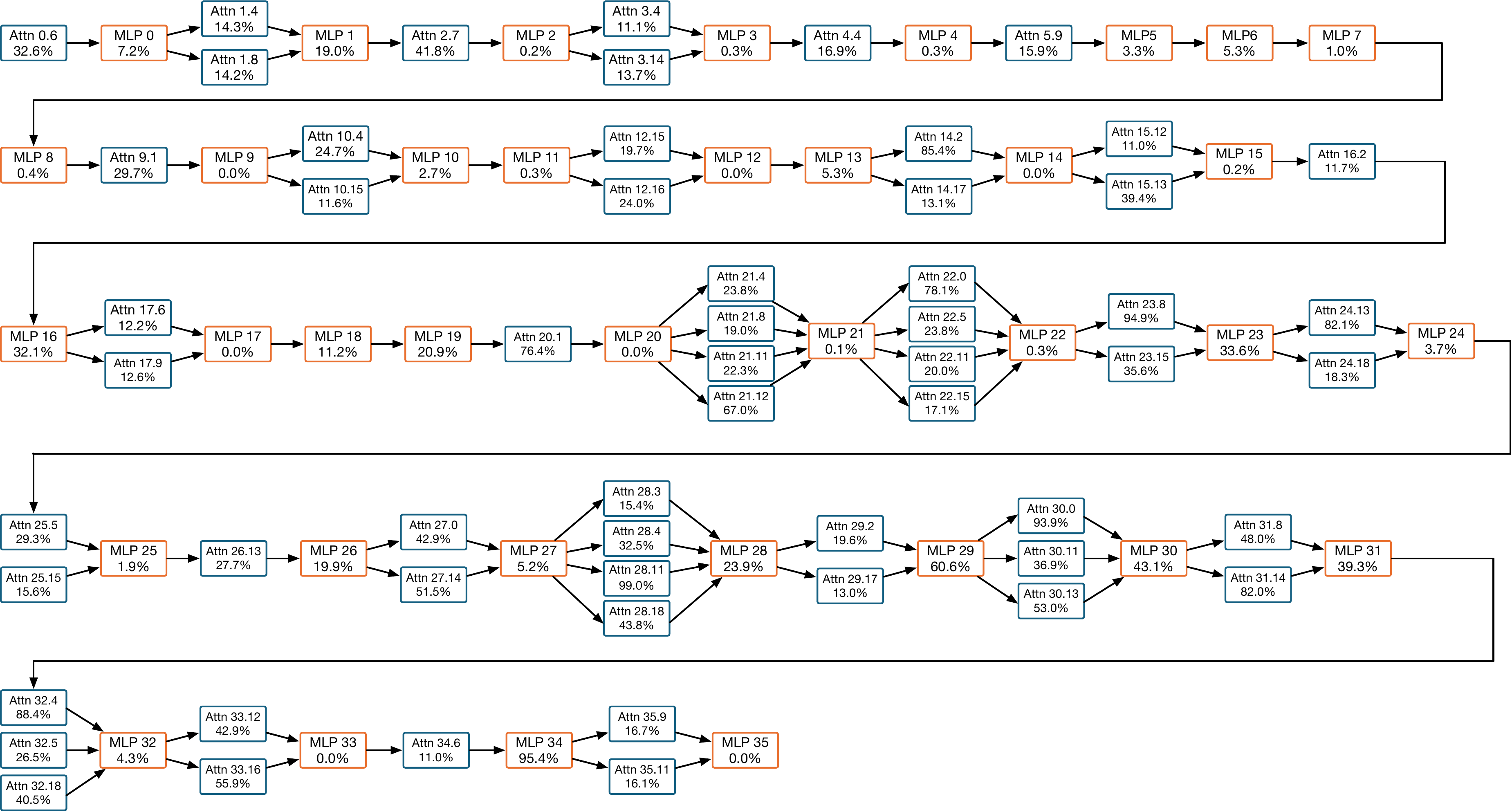}
    \caption{The full circuit of anchored bias for GPT2-Large model, where each attention head and MLP module are selected when MLP and attention pattern logit difference threshold is larger than 4. The percentage within each module indicates the probability of anchored bias across different datasets for GPT2-Large model when the threshold is larger than 4.}
    \label{fig:gpt2_large_circuit}
\end{figure*}

\begin{figure*}[ht]
    \centering
    \includegraphics[width=\textwidth]{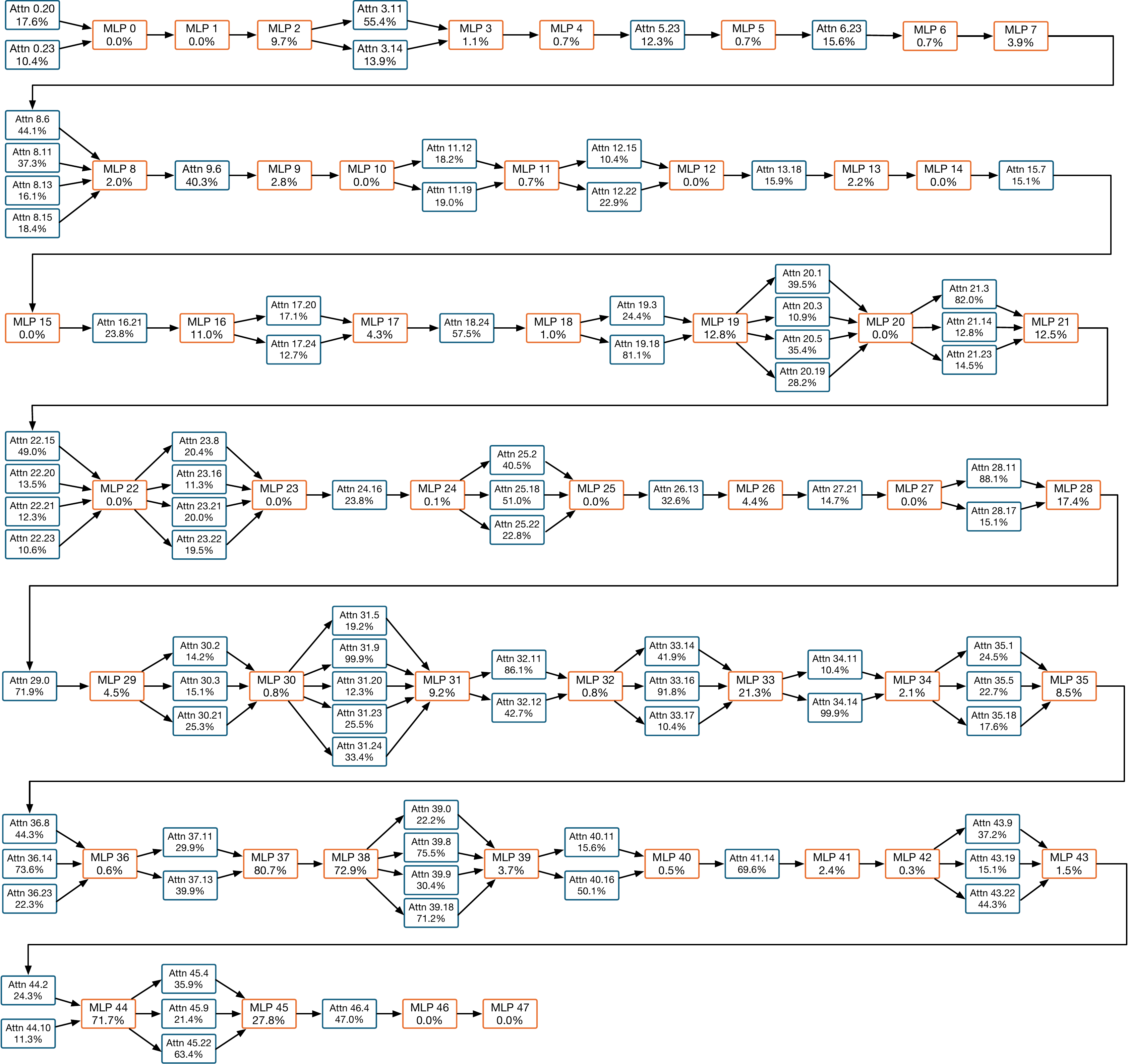}
    \caption{The full circuit of anchored bias for GPT2-XL model, where each attention head and MLP module are selected when MLP and attention pattern logit difference threshold is larger than 4. The percentage within each module indicates the probability of anchored bias across different datasets for GPT2-XL model when the threshold is larger than 4.}
    \label{fig:gpt2_xl_circuit}
\end{figure*}

Based on the identified MLP module and attention head using logit difference between anchored choice \texttt{`A'} and correct choices \texttt{`B,C,D,E'}. We can construct the full anchored-bias circuit for each GPT2 model. Fig.~\ref{fig:gpt2_medium_circuit},\ref{fig:gpt2_large_circuit},\ref{fig:gpt2_xl_circuit} show the full anchored-bias circuit of GPT2-Medium, GPT2-Large and GPT2-XL including identified attention heads and MLPs starting from layer 0 to the final layer.

\end{document}